\PassOptionsToPackage{table,dvipsnames}{xcolor}
\documentclass[]{fairmeta}
\usepackage[utf8]{inputenc}
\usepackage{url}
\usepackage{amsmath}
\usepackage{amsfonts}
\usepackage{nicefrac}
\usepackage{enumitem}
\usepackage{tabularx}
\usepackage{longtable}
\usepackage{wrapfig}
\usepackage{capt-of}
\usepackage{array}
\usepackage{pifont}
\usepackage{makecell}
\usepackage{listings}
\usepackage{placeins}
\usepackage{pdflscape}
\usepackage{float}
\usepackage{fontawesome5}

\newcommand{\rank}[1]{\hspace{0.35em}\raisebox{0.55ex}{\footnotesize\textcolor{blue!60!black}{#1}}}

\newcommand{\cmark}{\textcolor{green!45!black}{\ding{51}}}
\newcommand{\xmark}{\textcolor{red!70!black}{\ding{55}}}
\newcommand{\pmark}{\textcolor{orange!85!black}{$\triangle$}}
\newcolumntype{C}[1]{>{\centering\arraybackslash}m{#1}}
\newcolumntype{L}[1]{>{\raggedright\arraybackslash}p{#1}}
\newcolumntype{J}[1]{>{\hsize=#1\hsize\linewidth=\hsize\raggedright\arraybackslash}X}
\newcolumntype{K}[1]{>{\hsize=#1\hsize\linewidth=\hsize\centering\arraybackslash}X}
\newcolumntype{Y}{>{\centering\arraybackslash}X}

\newcommand{\spatialgenbench}{SpatialGen-Bench}

\definecolor{GroupBg}{gray}{0.93}

\setlabdisplayname{OmniAI Group of ZJU ACES Lab}
\setuniversityname{}

\title{Show, Don't Tell: Evaluating Spatial Cognition in Generative Pixels Rather Than LLM Text}

\author[*]{Xu Wang}
\author[*]{Kaixiang Yao}
\author{Miao Pan}
\author{Xiaohe Zhou}
\author{Xuanyu Liu}
\author[\ddagger]{Wenqi Zhang}
\author[\dagger\ddagger]{Xuhong Zhang}
\affiliation{Zhejiang University}
\contribution[*]{Equal contribution}
\contribution[\dagger]{Project lead}
\contribution[\ddagger]{Corresponding authors}
\metadata[\faGlobe\ Project]{\href{https://zju-omniai.github.io/ProVisE/}{\nolinkurl{https://zju-omniai.github.io/ProVisE/}}}
\metadata[\faGithub\ GitHub]{\href{https://github.com/ZJU-OmniAI/ProVisE}{\nolinkurl{https://github.com/ZJU-OmniAI/ProVisE}}}
\metadata[\faDatabase\ Hugging Face]{\href{https://huggingface.co/datasets/wx91726/SpatialGen-Bench}{\nolinkurl{https://huggingface.co/datasets/wx91726/SpatialGen-Bench}}}
\metadata[\faEnvelope\ Email]{\email{xuwang91726@gmail.com}; \email{zhangwenqi@zju.edu.cn}; \email{zhangxuhong@zju.edu.cn}}
\date{July 2026}

\hypersetup{
  pdftitle={Show, Don't Tell: Evaluating Spatial Cognition in Generative Pixels Rather Than LLM Text},
  pdfauthor={Xu Wang, Kaixiang Yao, Miao Pan, Xiaohe Zhou, Xuanyu Liu, Wenqi Zhang, Xuhong Zhang},
  pdfsubject={Protocolized visual evaluation of spatial cognition in image-generation models and vision-language models},
  pdfkeywords={spatial cognition, image generation, vision-language models, visual evaluation, benchmark, agentic protocol construction}
}

\abstract{%
Spatial intelligence is essential for agents to move from static semantic understanding toward interacting with the physical world. In the real world, many spatial tasks are grounded in continuous visual scenes, where locations, regions, and paths are more naturally expressed by pointing, marking, or drawing than by reporting precise coordinates or discrete textual symbols. However, existing spatial reasoning benchmarks usually require models to answer with coordinates, options, or textual descriptions, creating an answer-interface mismatch for image-generation models. This interface gap makes it difficult to evaluate image-generation models under the same task semantics as text-output VLMs, even though they can externalize spatial judgments directly in pixel space. To address this problem, we propose \textbf{ProVisE} (\textit{\textbf{Pro}tocolized \textbf{Vis}ual \textbf{E}valuation}), a benchmark-agnostic framework that elicits protocol-constrained visual answers from image-generation models and parses them into structured predictions compatible with the original evaluation metrics. ProVisE also includes an Agentic builder that constructs and validates task-specific protocols for new benchmarks. Building on this framework, we introduce \textbf{SpatialGen-Bench}, a curated diagnostic benchmark comprising 470 samples across 14 spatial subtasks, four capability levels, and diverse answer forms. We evaluate representative text-output VLMs and image-generation models in a unified setting, and validate Agentic protocol construction on six external spatial benchmarks. Experimental results show that image-generation models are competitive when spatial answers can be externalized directly in pixel space, while text-output VLMs retain a clear advantage in compositional spatial reasoning. These findings reveal the complementary strengths and limitations of pixel-space expression and text-based reasoning, while ProVisE and SpatialGen-Bench establish a metric-compatible testbed for generative spatial evaluation and future studies of spatial cognition in image-generation models.
}

\begin{document}

\maketitle
\thispagestyle{firstheader}

\section{Introduction}

Spatial intelligence is a foundational prerequisite for artificial agents to move beyond static semantic understanding and interact safely with the physical world~\citep{yu2025far, liu2026spatial, zheng2025multimodal}.
It spans a broad range of perceptual and reasoning abilities, including visual grounding, affordance prediction, target-region localization, and relational spatial reasoning over positions, distances, and orientations.
As vision-language models (VLMs)~\citep{achiam2023gpt, comanici2025gemini, bai2025qwen3vltechnicalreport} have become increasingly capable, substantial effort has been devoted to improving their spatial perception and reasoning abilities~\citep{chen2024spatialvlm, wu2025spatial, cheng2024spatialrgpt,cai2025spatialbot, liu2025ssr, zhou2025roborefer, yang2025cambrian, li2025spatialladder}.
Following progress in visual question answering (VQA) and spatial VLM evaluation~\citep{yang2025thinking, fu2024blink, wang2026mindcubespatialmentalmodeling, yang2025mmsi,ray2024sat,jia2025omnispatial,li2025viewspatial,tong2024cambrian,du2024embspatial,zhang2025spinbench}, existing spatial intelligence research has largely been framed around text-output VLMs.
Models are typically prompted and evaluated through symbolic or structured answer formats, such as coordinates, option labels, or textual descriptions.
This paradigm makes spatial reasoning measurable for VLMs, but it also forces continuous visual judgments into textual or coordinate-based outputs, which are often unnatural for spatial tasks.

Recent progress in generative vision models motivates a closer evaluation of their spatial cognition.
Unlike text-output VLMs, image-generation models operate directly in pixel space and can express spatial judgments by modifying, marking, or synthesizing visual content.
This makes them natural candidates for spatial tasks whose answers are continuous or visually grounded, rather than purely symbolic.
Recent work shows that image generators can produce decodable visualizations for dense vision tasks such as segmentation, depth estimation, and surface-normal prediction~\citep{gabeur2026image}, while video-generation models exhibit zero-shot behaviors across perception, manipulation, and early visual reasoning~\citep{wiedemer2025videomodelszeroshotlearners}.
Related progress in native and unified multimodal generation further suggests that visual generation may encode representations useful for understanding and world modeling~\citep{cui2025emu3, xie2025show, deng2025emerging}.
These studies suggest that generative vision models are moving beyond image synthesis, yet their spatial cognition remains largely unexplored when spatial answers can be expressed visually.

Existing evaluation lines leave this gap unresolved.
Text-to-image benchmarks assess prompt fidelity, compositional alignment, or spatial relations in synthesized images~\citep{hu2023tifa, ghosh2023geneval, huang2025t2i, li2024genai, bakr2023hrsbenchholisticreliablescalable, gokhale2022benchmarking, wang2026everything}; they evaluate whether a generated image satisfies a text prompt, not whether a model can answer a spatial question grounded in an input scene.
Conversely, spatial reasoning benchmarks for VLMs focus on text-output models and score symbolic or structured predictions~\citep{liu2023visual, wu2025spatialscore, xu2026spatialbenchbenchmarkingmultimodallarge, stogiannidis2025mindgapbenchmarkingspatial, Wu_2025_ICCV}.
What remains missing is a metric-compatible way to compare image-generation models and text-output VLMs under shared spatial task semantics.

\begin{figure}[t]
  \centering
  \includegraphics[width=\textwidth]{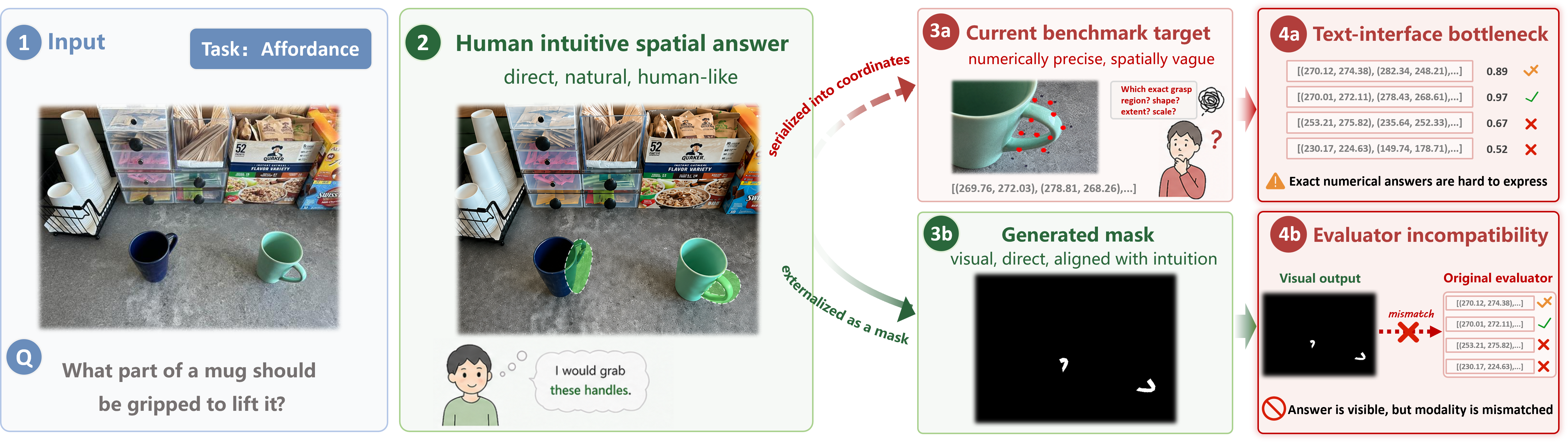}
  \caption{\textbf{The answer-interface mismatch in spatial evaluation.}
  A spatial judgment can be expressed directly as a graspable region, whereas existing benchmarks often require an exact coordinate list. This burdens text interfaces with precise serialization and leaves a generated visual answer incompatible with the original evaluator.}
  \label{fig:motivation}
\end{figure}

The main obstacle is the answer interface, as illustrated in Figure~\ref{fig:motivation}.
Existing spatial benchmarks are primarily designed for text-output VLMs and typically require coordinates, option labels, or textual descriptions as final answers~\citep{wu2025spatialscore}.
For tasks involving regions, paths, depths, or orientations, this requires models to express spatial judgments as numerical or symbolic outputs, such as bounding-box coordinates $[x_1, y_1, x_2, y_2]$, per-pixel depth values, or direction labels.
This is analogous to asking a person to report exact pixel coordinates instead of pointing to or sketching a region.
Such an interface is unnatural for text-output VLMs, which must verbalize spatial structure, and even more restrictive for image-generation models, whose native responses are visual.
Although an image-generation model may mark an object, highlight a region, or draw a trajectory, these responses cannot be directly scored by benchmarks expecting textual or coordinate-based predictions.
As a result, their spatial cognition cannot be systematically compared with text-output VLMs under the original task metrics.

A natural workaround is to use an auxiliary VLM judge to read the generated images and infer the final answer~\citep{wu2025spatialscore, wang2026everything}.
However, judge-based scores depend on the judge model's own reliability and may reflect judge uncertainty rather than the generated spatial answer itself~\citep{cho2024davidsonianscenegraphimproving}.
They are also difficult to align with the original metrics used for text-output VLMs, making controlled comparison between VLMs and image-generation models unreliable under shared task semantics.

To make generative spatial cognition measurable, we introduce an evaluation framework together with a diagnostic benchmark.
\textbf{ProVisE} (\textit{\textbf{Pro}tocolized \textbf{Vis}ual \textbf{E}valuation}) is a benchmark-agnostic framework that elicits protocol-constrained visual answers from image-generation models.
Instead of requiring textual or coordinate-based responses, ProVisE constrains generated outputs into predefined visual formats and parses them into structured predictions compatible with the original task metrics.
Because manually engineering a protocol for every new task does not scale, ProVisE also provides an Agentic construction path. It normalizes a new benchmark into task-level contracts, constructs an appropriate generation--parser protocol for each task, and validates the resulting artifact before model evaluation.
We further construct \textbf{SpatialGen-Bench}, a curated diagnostic benchmark for spatial cognition in image-generation models.
It covers 14 subtasks across perception, understanding, reasoning, and interaction, providing a capability-level view of model strengths and weaknesses.
Its diverse answer forms also allow us to examine whether ProVisE can support metric-compatible visual evaluation across different answer interfaces.
Together, ProVisE and SpatialGen-Bench enable systematic comparison between text-output VLMs and image-generation models under shared task semantics.

The suite is designed to be reusable beyond a single benchmark or model family.
It can be applied to contemporary image-generation models such as FLUX, Seedream, and Janus~\citep{labs2025flux, seedream2025seedream, chen2025janus}, while preserving compatibility with VLM-oriented spatial benchmarks such as MindCube, SAT, and VSR~\citep{wang2026mindcubespatialmentalmodeling, ray2024sat, liu2023visual}.
By converting visual responses back into metric-compatible predictions, ProVisE enables cross-model and cross-benchmark evaluation without treating a post-hoc VLM judge as the default evaluator.

By deploying this evaluation suite, we provide a systematic comparison of spatial cognition in image-generation models and text-output VLMs.
Image-generation models are most competitive when answers can be externalized as inspectable pixel-space evidence, while text-output VLMs remain stronger when tasks require compositional reasoning and precise spatial transformations.
Additional experiments show that universal VLM parsing changes both scores and rankings, while most visual failures arise from incorrect spatial predictions rather than generation, protocol, or parser breakdowns.
Experiments on six external benchmarks further show that the Agentic builder can adapt ProVisE to heterogeneous task and metric contracts.
Together, these results reveal complementary strengths in visual externalization and text-based reasoning and motivate their coordinated use.

\FloatBarrier
In summary, our main contributions are as follows:  

\begin{itemize}

    \item \textbf{Modality Mismatch in Generative Spatial Evaluation}:
    We investigate the spatial cognition of image-generation models and reveal an answer-interface mismatch in existing spatial reasoning benchmarks: reducing spatial judgments to textual or coordinate-based formats can obscure the assessment of text-output VLMs and prevent image-generation models from being evaluated under the same task semantics and metrics.

    \item \textbf{A Unified Suite for Image-Generation Spatial Evaluation}: 
    We propose ProVisE, a benchmark-agnostic framework that enables image-generation models to be evaluated in pixel space while preserving compatibility with the original benchmark metrics. In addition, we introduce SpatialGen-Bench, a curated diagnostic benchmark with 14 subtasks across four spatial capability levels and diverse answer forms.

    \item \textbf{Agentic Protocol Construction}:
    We introduce an Agentic builder that constructs and validates task-specific generation--parser protocols for previously unconfigured benchmarks. Each resulting protocol is shared across target models, extending ProVisE beyond manually configured tasks.

    \item \textbf{Systematic Comparison of Generative Models and VLMs}: 
    Through extensive experiments, we provide a quantitative profile of how image-generation models and text-output VLMs differ across spatial capabilities. The results reveal complementary strengths: image-generation models are more competitive when answers can be externalized directly in pixel space, while text-output VLMs remain stronger in compositional spatial reasoning.

\end{itemize}

\section{Protocolized Visual Evaluation}

\begin{figure}[t]
  \centering
  \includegraphics[width=0.98\textwidth]{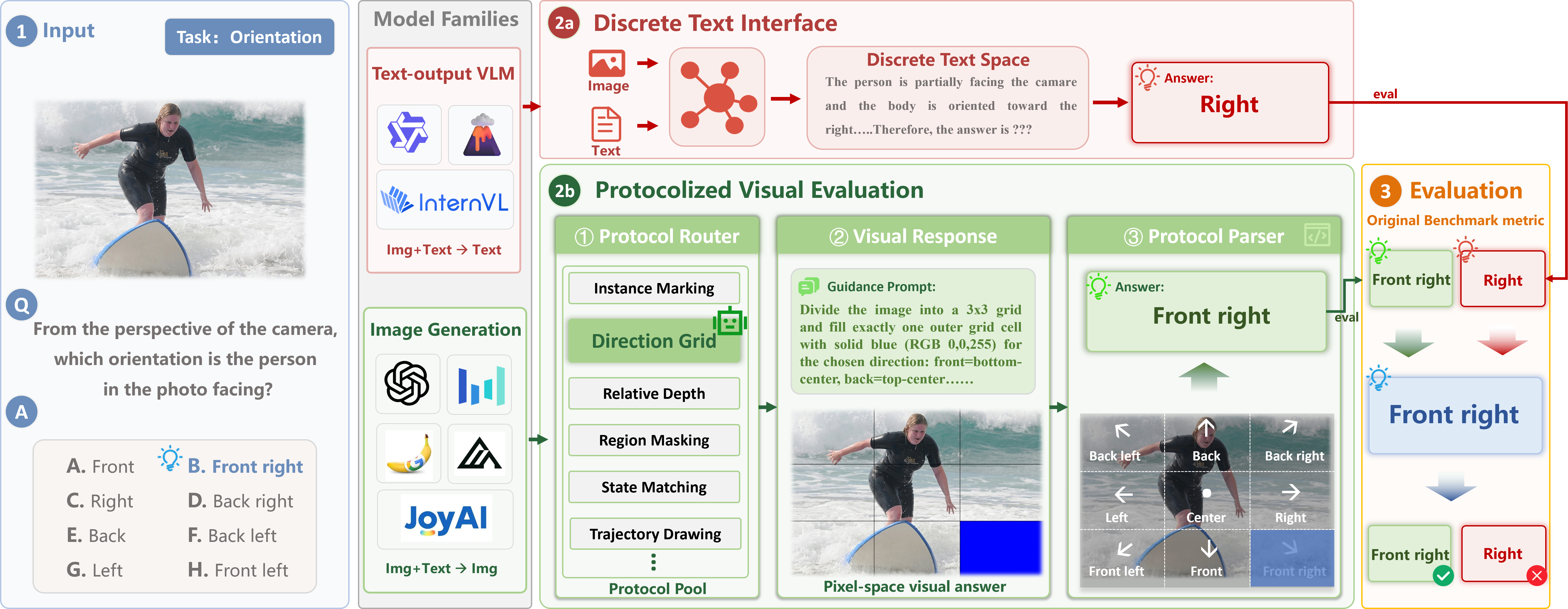}
  \caption{\textbf{Overview of ProVisE.}
  Text-output VLMs follow \textbf{2a} and are scored directly by the original evaluator.
  Image-generation models follow \textbf{2b}, where a routed protocol guides generation and parsing into structured answers.
  Both are scored by the same source-task metric under shared semantics.}
  \label{fig:protocolized_workflow}
\end{figure}

This section introduces the ProVisE pipeline for adapting image-generation models to spatial benchmarks designed for textual answers, as shown in Figure~\ref{fig:protocolized_workflow}. The pipeline preserves the source-task metric while replacing only the response interface with a constrained visual protocol. We first define visual protocols and routing, then describe Agentic construction, protocol execution and parsing, and metric-compatible evaluation.

\subsection{Visual Protocols and Routing}
\label{sec:protocol-routing}

A visual protocol is the basic unit of ProVisE. It pairs a guidance prompt, which constrains how a model expresses its answer in pixel space, with a parser that converts the generated image into a structured prediction. The protocol also specifies the expected answer format, invalid-output conditions, and its mapping to the original task metric. Because these elements are fixed before model evaluation, the generated response can be interpreted without subjective post-hoc decisions.

To cover different spatial tasks, we construct a task-aware protocol pool in which every visual answer format is paired with a corresponding parser.
The pool spans instance and point marking, direction grids, relative-depth maps, region masks, state matching, trajectory drawing, and task-grounded evidence rendering; Appendix~\ref{app:parser_rules} gives the operational parser rules and invalid-output boundaries.

Protocol routing determines one protocol for each task before target-model evaluation. If a benchmark already has a validated configuration, ProVisE uses it directly. Otherwise, the Agentic builder constructs a task-specific protocol from the benchmark's input, answer, and metric contract. The accepted configuration is then shared by all evaluated models, separating protocol design from model comparison.

\subsection{Agentic Protocol Construction}
\label{sec:agentic-construction}

\begin{figure}[t]
  \centering
  \includegraphics[width=0.98\textwidth]{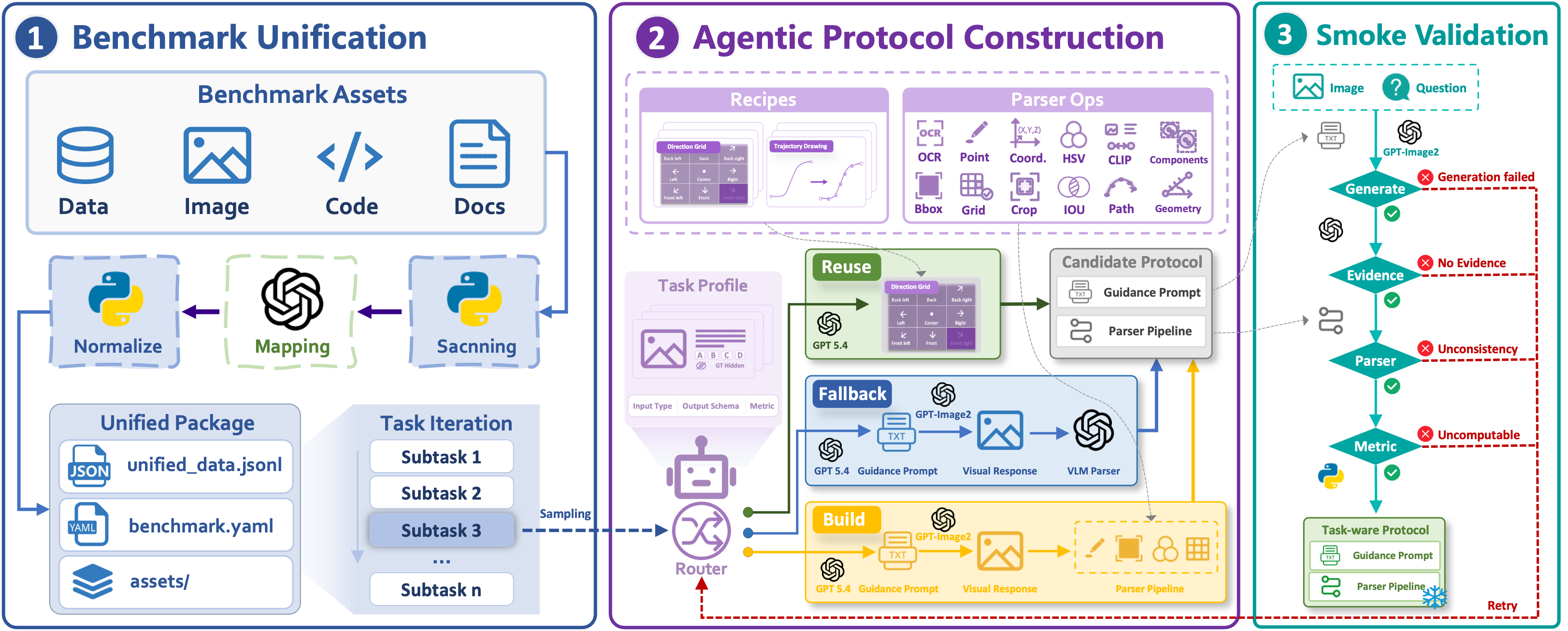}
  \caption{\textbf{Agentic protocol construction in ProVisE.}
  A benchmark is normalized into task contracts; the Agent constructs a protocol for each task; and the resulting generation--parser pair is automatically validated before being applied unchanged to every target model.}
  \label{fig:agentic-construction}
\end{figure}

For a previously unconfigured benchmark, ProVisE proceeds in the three stages summarized in Figure~\ref{fig:agentic-construction}. First, it identifies the benchmark records, input media, task labels, answer formats, and evaluation metrics, then normalizes them into task-level contracts without changing the task semantics or scoring rules.

Second, the Agent examines representative examples from each task and selects one of three construction routes. \textbf{Reuse} selects an existing compatible protocol, \textbf{Build} assembles a task-specific protocol from registered parsing components, and \textbf{Fallback} uses a constrained auxiliary parser when the available deterministic components cannot recover the required answer form. Every route produces the same declarative artifact: a generation prompt, parser configuration, answer contract, invalid-output rules, and metric mapping. The Agent does not generate executable evaluation code.

Finally, ProVisE tests each candidate on a small set of examples. Smoke validation checks successful visual generation, stable parsing, and compatibility with the original metric. A failed protocol may be revised once before rejection or transition to Fallback. Accepted configurations are then applied unchanged to all target models, preventing model-specific protocol optimization. Appendix~\ref{app:agentic-construction-coverage} reports the normalization fields, route definitions, validation criteria, and task-level outcomes.

\subsection{Protocol Execution and Parsing}
\label{sec:prompt-parser-execution}

After protocol routing or construction, ProVisE applies the selected protocol to guide visual generation, as illustrated by the 2b branch in Figure~\ref{fig:protocolized_workflow}. The image-generation model is instructed to produce a visual response under the structural constraints of the protocol, rather than a free-form image. Depending on the task, the response may take the form of marked instances, a direction grid, a binary region mask, a grounded point, a rendered spatial state, or a drawn trajectory. The corresponding parser then converts that response into the structured answer required by the original benchmark. Deterministic parsing is preferred. When Fallback is necessary, the parser receives only the generated response, the fixed visual-evidence contract, the required answer format, and candidate choices; it does not receive the source image or original question and therefore cannot independently re-solve the benchmark item.

\begin{figure}[H]
  \centering
  \includegraphics[width=\textwidth]{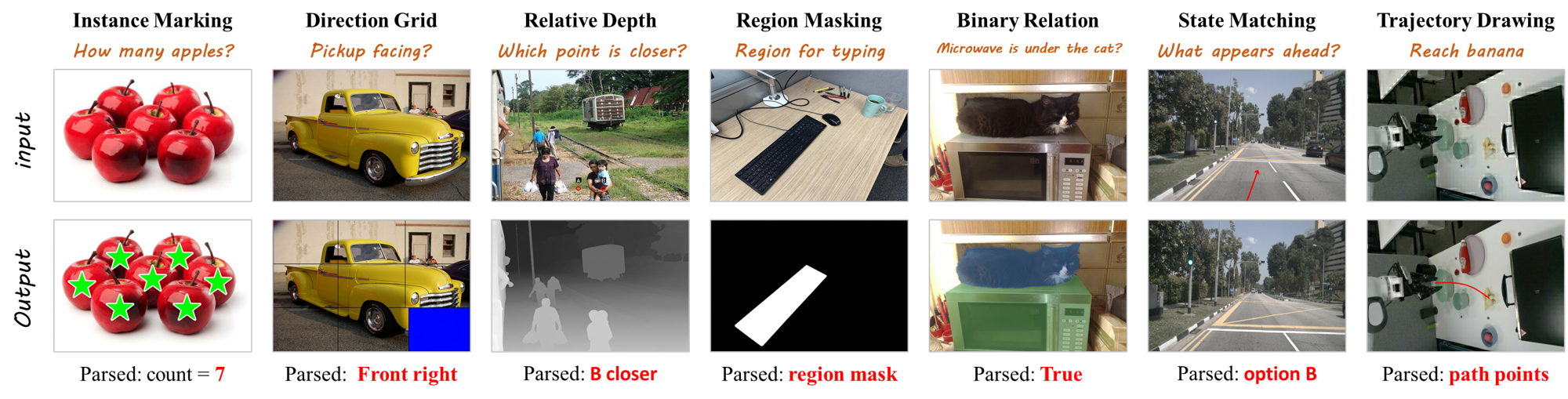}
  \caption{Representative protocol-execution cases for ProVisE. Each column shows a benchmark input (top), the protocolized output (bottom), and the recovered structured prediction.}
  \label{fig:protocol_examples}
\end{figure}

The protocol-constrained response provides an explicit intermediate representation of the model's answer. As shown in Figure~\ref{fig:protocol_examples}, each column pairs a benchmark input with the visual response generated under a specific protocol. These examples illustrate how different answer forms can be expressed visually while still being parsed into standardized predictions.

\subsection{Metric-Compatible Evaluation}
\label{sec:metric-compatible-evaluation}

After parsing, each visual response is converted into the answer space expected by the original benchmark, such as a label, count, mask, candidate state, or trajectory. The resulting prediction is then scored using the original task metric.
For text-output VLMs, raw responses are directly normalized to the required answer format; for image-generation models, visual responses are parsed into that format after protocol-guided generation. Both outputs are therefore evaluated by the same source-task metric, corresponding to the evaluation stage in Figure~\ref{fig:protocolized_workflow}. In this way, ProVisE changes only the response interface for image-generation models while preserving the task target and scoring rule. Most protocols use deterministic image processing, geometry, or fixed similarity models. Fallback tasks use a protocol-constrained generated-image-only parser, which we distinguish from unconstrained post-hoc VLM judging in Sec.~\ref{sec:vlm-judge-analysis}.

\FloatBarrier
\section{SpatialGen-Bench}\label{sec:spatialgen-bench}

To examine ProVisE across heterogeneous spatial answer interfaces, we construct SpatialGen-Bench with tasks spanning discrete decisions, counts, regions, states, and trajectories. The benchmark preserves source-task semantics and metrics while organizing spatial cognition across perception, understanding, reasoning, and interaction, providing the basis for our diagnostic analysis.

\subsection{Capability Hierarchy}
\label{sec:capability-hierarchy}

\begin{figure}[t]
  \centering
  \includegraphics[width=\textwidth]{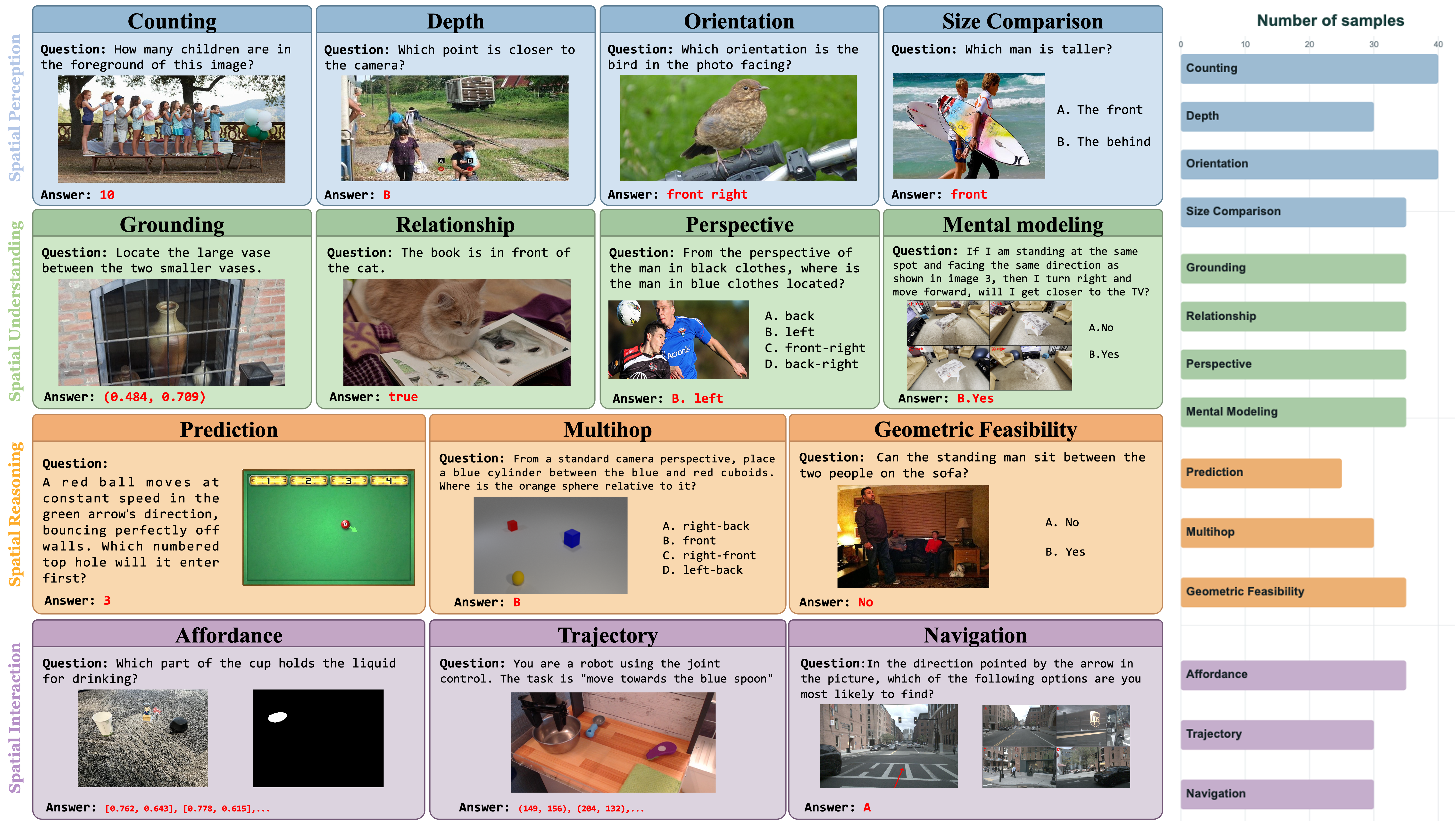}
  \caption{\textbf{Overview of \spatialgenbench{}.} Tasks are organized into \textcolor{RoyalBlue}{perception}, \textcolor{ForestGreen}{understanding}, \textcolor{BurntOrange}{reasoning}, and \textcolor{Purple}{interaction}.}
  \label{fig:spatialgen-overview}
\end{figure}

SpatialGen-Bench organizes spatial cognition into four progressive levels, as shown in Figure~\ref{fig:spatialgen-overview}. 
The hierarchy starts from basic spatial attribute recognition, to viewpoint understanding and scene layout comprehension, to complex spatial reasoning, and ultimately to action-oriented decision making.

\textbf{Spatial perception} forms the foundation of the hierarchy, covering basic spatial properties grounded in direct visual evidence: (1) counting, (2) relative depth estimation, (3) orientation judgment, and (4) object size comparison. These tasks contain 145 samples in total.

\textbf{Spatial understanding} moves beyond local cues toward scene-level representation. 
It includes (5) relationship verification, (6) perspective judgment, (7) spatial mental modeling, and (8) spatial grounding, which require the model to connect objects, adjust viewpoints, infer layouts, and localize a referred target. These tasks contain 140 samples.

\textbf{Spatial reasoning} builds on scene representation to infer changes or conclusions that are not directly specified. 
It includes (9) multi-hop spatial reasoning, (10) spatial state prediction, and (11) geometric feasibility, covering compositional relations, physically plausible state changes, and fit or collision constraints. These tasks contain 90 samples.

\textbf{Spatial interaction} represents the transition from spatial understanding to action in the real world and includes (12) affordance grounding, (13) navigation, and (14) trajectory planning, where the model must decide where interaction is possible, which state to move toward, and how to reach the target. These tasks contain 95 samples.

This layered design supports broad diagnostic evaluation from static perception to action-oriented interaction and provides an interpretable basis for identifying which capability levels remain challenging for current models.
To clarify the positioning of SpatialGen-Bench, Table~\ref{tab:benchmark_comparison} compares it with other spatial evaluation benchmarks.
While prior benchmarks cover important spatial settings such as perception, localization, mental modeling, reasoning, and text-to-image generation, SpatialGen-Bench jointly supports hierarchical capability diagnosis, multiple answer forms, and shared-semantics evaluation of text-output VLMs and image-output generative models.

\subsection{Benchmark Construction and Evaluation}

\noindent\textbf{Data sources.}
SpatialGen-Bench draws from multiple public spatial evaluation benchmarks~\citep{paiss2023teaching,fu2024blink,liu2023visual,jung2025right,li2025viewspatial,wu2026visualgenerationunlockshumanlike,wang2026mindcubespatialmentalmodeling,zhang2025sphere,mao2016generation,hao2025roboafford++,RoboBrain1.0,chow2025physbenchbenchmarkingenhancingvisionlanguage}. The selected tasks cover object-level perception, relational and viewpoint reasoning, object size, spatial grounding, state modeling, geometric feasibility, affordance grounding, navigation, and path planning. Their interfaces range from discrete choices and counts to points, masks, candidate states, and continuous trajectories, providing the heterogeneous answer contracts required by ProVisE.

\noindent\textbf{Quality-aware sample selection.}
For each selected task, we collect candidate samples from the corresponding source data and manually inspect them before inclusion.
Samples are retained only when the input media, original annotations, task instruction, and answer can be reliably recovered and scored under the source-task semantics.
We remove near-duplicates, visually unclear cases, inconsistent annotations, and samples with ambiguous or multiple valid answers.

\noindent\textbf{Spatial capability mapping.}
After filtering, each instance is assigned to a subtask and capability level according to its objective and required spatial competence rather than its source benchmark; the assignment is checked against the source annotation and task formulation. The resulting samples use standardized shared fields while retaining task-specific scoring metadata, as detailed in Appendix~\ref{app:unified_instance_format}.

\noindent\textbf{Data integrity and review.}
Every released item undergoes deterministic asset and schema checks plus manual inclusion review. Appendix~\ref{app:source_data_quality_assessment} documents the checked fields and review coverage.

For evaluation, counting and most discrete-answer subtasks use exact-match accuracy after answer normalization, including object size and geometric feasibility. Spatial grounding uses point-in-mask accuracy. Affordance uses mean mask precision, prediction uses the partial-credit endpoint score defined in Appendix~\ref{app:score_aggregation}, and trajectory planning uses DFD-success@0.4 based on Discrete Fr\'echet Distance.

\begin{table}[t]
\centering
\footnotesize
\setlength{\tabcolsep}{1.8pt}
\renewcommand{\arraystretch}{1.10}
\setlength{\aboverulesep}{1.5ex}
\setlength{\belowrulesep}{1.5ex}
\caption{
Comparison with representative spatial evaluation benchmarks.
}
\label{tab:benchmark_comparison}
\begin{tabular}{@{}
>{\raggedright\arraybackslash}m{0.175\textwidth}
>{\raggedright\arraybackslash}m{0.185\textwidth}
>{\centering\arraybackslash}m{0.14\textwidth}
>{\raggedright\arraybackslash}m{0.195\textwidth}
>{\centering\arraybackslash}m{0.075\textwidth}
>{\centering\arraybackslash}m{0.065\textwidth}
>{\centering\arraybackslash}m{0.065\textwidth}
@{}}
\toprule
\textbf{Benchmark} &
\textbf{Main scope} &
\textbf{Scale} &
\textbf{Task organization} &
\textbf{Hierarchy} &
\shortstack[c]{\textbf{Ans.}\\[-0.35ex]\textbf{types}} &
\shortstack[c]{\textbf{Shared}\\[-0.35ex]\textbf{eval.}} \\
\midrule

BLINK~\citep{fu2024blink}
& Core visual perception
& 3.8K MCQs; 14 tasks
& Perception-oriented task suite
& \xmark
& \xmark
& \xmark \\

ViewSpatial-Bench~\citep{li2025viewspatial}
& Multi-perspective spatial localization
& 5.7K+ QA pairs; 5 task types
& Camera- and person-centric five-task taxonomy
& \pmark
& \xmark
& \xmark \\

MindCube~\citep{wang2026mindcubespatialmentalmodeling}
& Spatial mental modeling from limited views
& 21K+ questions; 3.2K+ images
& Cognitive mapping, perspective taking, and mental simulation
& \xmark
& \pmark
& \xmark \\

OmniSpatial~\citep{jia2025omnispatial}
& Comprehensive spatial reasoning for VLMs
& 8K+ QA pairs; 50 subcategories
& 4 spatial categories with 50 subcategories
& \cmark
& \xmark
& \xmark \\

SpatialScore~\citep{wu2025spatialscore}
& Comprehensive multimodal spatial intelligence
& 5K verified samples; 30 tasks
& Spatial tasks across data types and QA formats
& \cmark
& \pmark
& \xmark \\

SpatialTree~\citep{spatialtree2025}
& Hierarchical spatial ability evaluation
& 6.3K records; 11 tasks
& Four-level capability tree
& \cmark
& \pmark
& \xmark \\

SpatialGenEval~\citep{wang2026everything}
& Spatial reasoning in text-to-image generation
& 1,230 prompts; 12,300 MCQs
& Spatial domains with 10 sub-domains
& \cmark
& \xmark
& \pmark \\

\textbf{SpatialGen-Bench}
& \textbf{Spatial cognition benchmark for generative evaluation}
& \textbf{470 samples; 14 subtasks}
& \textbf{Four capability levels with curated subtasks}
& \cmark
& \cmark
& \cmark \\

\bottomrule
\end{tabular}

\begin{minipage}{0.98\textwidth}
\footnotesize
\cmark: full support; \pmark: partial support; \xmark: not a design target.
\textit{Hierarchy}: hierarchical capability diagnosis; \textit{Ans. types}: multiple evaluated answer forms; \textit{Shared eval.}: text-output VLMs and image-generation models evaluated under shared task semantics and metrics.
\end{minipage}
\end{table}

\FloatBarrier
\section{Experiments}

We use \spatialgenbench{} as a diagnostic testbed for evaluating spatial cognition through protocolized visual answers.
The experiments compare text-output VLMs and image-generation models under shared task semantics, and examine whether ProVisE provides a reliable alternative to universal black-box parsing.
We focus on five questions:
\textbf{(1)} which task properties distinguish visual from textual spatial answering;
\textbf{(2)} whether the two interfaces solve the same examples or expose complementary successes;
\textbf{(3)} how sensitive scores and rankings are to replacing task-specific parsers with one universal VLM parser;
\textbf{(4)} whether failed visual evaluations arise from generation or parsing breakdowns, or from incorrect spatial predictions;
and \textbf{(5)} whether ProVisE supports executable evaluation on heterogeneous external spatial benchmarks.

\subsection{Models and Evaluation Settings}

We report 31 model-interface systems, grouped by answer interface: 20 \textbf{Text Answering} systems and 11 \textbf{Visual Answering} systems. They cover general-purpose, spatially specialized, embodied, unified, and dedicated image-generation models. When a model supports both interfaces, each is treated as a separate system because it exposes a different answer mechanism.

Each task's visual protocol is frozen before target-model evaluation. Its generation prompt, parser, invalid-output rules, and metric mapping are shared by all visual-answering models, with no model-specific tuning. Text systems answer in the original answer space. Missing, unreadable, malformed, or unparseable outputs remain in the common denominator and receive zero.

Source-task metrics produce per-sample scores in $[0,1]$; scores are averaged within each task and macro-averaged across tasks for capability and overall results. Appendix~\ref{app:evaluation_settings} reports the complete metrics, task-level scores, runtime settings, and bootstrap uncertainty.

\FloatBarrier
\subsection{Main Results on \spatialgenbench{}}
\label{sec:main-results}

\begin{table}[H]
\centering
\footnotesize
\setlength{\tabcolsep}{5.5pt}
\renewcommand{\arraystretch}{1.00}
\caption{
Main results on SpatialGen-Bench (\%). Capability scores macro-average their constituent tasks, and Overall macro-averages task scores. Bold marks the best model within each answer interface.
}
\label{tab:main-results-vertical}
\begin{tabular}{@{\hspace{6pt}}l*{5}{c}@{\hspace{6pt}}}
\toprule
\textbf{Model} & \textbf{Perception} & \textbf{Understanding} & \textbf{Reasoning} & \textbf{Interaction} & \textbf{Overall} \\
\midrule
\rowcolor{black!10}
\textbf{Human} & 97.41 & 85.71 & 77.92 & 87.60 & 87.79 \\
\midrule
\rowcolor{blue!5}
\multicolumn{6}{l}{\textbf{\textit{Text Answering}}} \\
GPT-5.4 & 74.61 & 44.29 & 56.64 & \textbf{69.70} & \textbf{61.04} \\
Cosmos3-Nano & \textbf{76.82} & 45.00 & 55.20 & 63.15 & 60.17 \\
RynnBrain-8B & 67.05 & \textbf{45.71} & 44.21 & 44.13 & 51.15 \\
RoboBrain2.5-8B-NV & 58.39 & 34.29 & 47.76 & 50.03 & 47.43 \\
SenseNova-Vision-7B-MoT & 54.40 & 43.57 & 40.85 & 46.83 & 46.78 \\
InternVL3.5-8B & 51.70 & 36.43 & \textbf{62.45} & 35.87 & 46.25 \\
SpaceR & 61.04 & 31.43 & 53.24 & 37.38 & 45.84 \\
SpaceOm & 59.23 & 40.00 & 43.78 & 32.78 & 44.76 \\
LLaVA-OV-1.5-8B & 54.23 & 37.86 & 39.97 & 45.56 & 44.64 \\
SpatialThinker-3B & 57.65 & 41.43 & 39.75 & 36.37 & 44.62 \\
Qwen3-VL-8B & 59.79 & 37.14 & 47.00 & 31.43 & 44.50 \\
GEM-2B & 56.70 & 43.57 & 39.70 & 30.32 & 43.65 \\
Cambrian-S-7B & 68.01 & 21.43 & 47.09 & 32.22 & 42.55 \\
GLM-4.6V-Flash & 56.28 & \textbf{45.71} & 29.75 & 25.42 & 40.96 \\
VLM-3R-7B & 56.19 & 21.43 & 37.52 & 31.21 & 36.90 \\
Spatial-MLLM & 32.77 & 30.00 & 32.97 & 36.51 & 32.82 \\
SpatialRGPT-8B & 38.90 & 27.86 & 29.47 & 29.52 & 31.72 \\
Janus-Pro-7B & 39.20 & 26.43 & 35.75 & 24.13 & 31.58 \\
Janus-1.3B & 39.43 & 28.57 & 31.62 & 11.11 & 28.59 \\
SpatialBot-3B & 33.12 & 32.86 & 31.85 & 10.79 & 27.99 \\
\midrule
\rowcolor{green!5}
\multicolumn{6}{l}{\textbf{\textit{Visual Answering}}} \\
GPT Image 2 & 69.58 & \textbf{47.14} & \textbf{41.83} & \textbf{56.80} & \textbf{54.49} \\
Nano Banana 2 & \textbf{72.92} & 42.14 & 24.92 & 48.61 & 48.63 \\
GPT-5 Image Mini & 64.40 & 33.57 & 26.89 & 29.13 & 40.00 \\
Seedream 4.5 & 51.43 & 35.71 & 28.69 & 33.04 & 38.13 \\
JoyAI-Image & 41.90 & 35.00 & 19.50 & 42.61 & 35.28 \\
FLUX.2 [klein] 9B & 47.68 & 26.43 & 22.80 & 35.09 & 33.58 \\
SenseNova-Vision-7B-MoT & 32.44 & 45.71 & 12.44 & 35.93 & 32.70 \\
FLUX.2 [klein] 4B & 41.93 & 27.14 & 14.11 & 28.37 & 28.84 \\
OmniGen-v1 & 29.85 & 25.00 & 26.36 & 25.21 & 26.72 \\
Qwen-Image-Edit-2511 & 30.65 & 24.29 & 29.69 & 20.47 & 26.44 \\
BAGEL-7B-MoT & 27.38 & 23.57 & 24.44 & 26.32 & 25.44 \\
\bottomrule
\end{tabular}
\end{table}

\textbf{Current models remain substantially behind humans.}
As summarized in Table~\ref{tab:main-results-vertical}, GPT-5.4 leads text answering with an Overall score of 61.04, while GPT Image 2 leads visual answering with 54.49; both remain well below the Human reference at 87.79. Relative to the best machine score in each column, the largest capability-level gap is in Understanding (38.57 points), and the widest task-level gaps occur on Perspective (51.43), Relationship (31.43), and Mental Modeling (28.57), as detailed in Table~\ref{tab:full-task-results}. The principal bottleneck is therefore maintaining relation semantics and transforming spatial representations across viewpoints, rather than generic multi-step reasoning alone.

\textbf{Spatially specialized models still fall short of general spatial intelligence.}
SpaceR attains the strongest machine Relationship score (57.14), while RynnBrain-8B and RoboBrain2.5-8B perform particularly well on Grounding and Affordance; RoboBrain2.5-8B also reaches 88.57 on Size. However, their performance varies sharply across Orientation, Perspective, Navigation, and Trajectory. Under the broad task coverage of SpatialGen-Bench, these uneven profiles show that current spatial specialization improves selected geometric or action-grounding skills but does not yet yield general spatial competence.

\subsection{Interface Strengths and Complementarity}
\label{sec:interface-strengths}

\begin{figure}[H]
    \centering
    \includegraphics[width=0.98\textwidth]{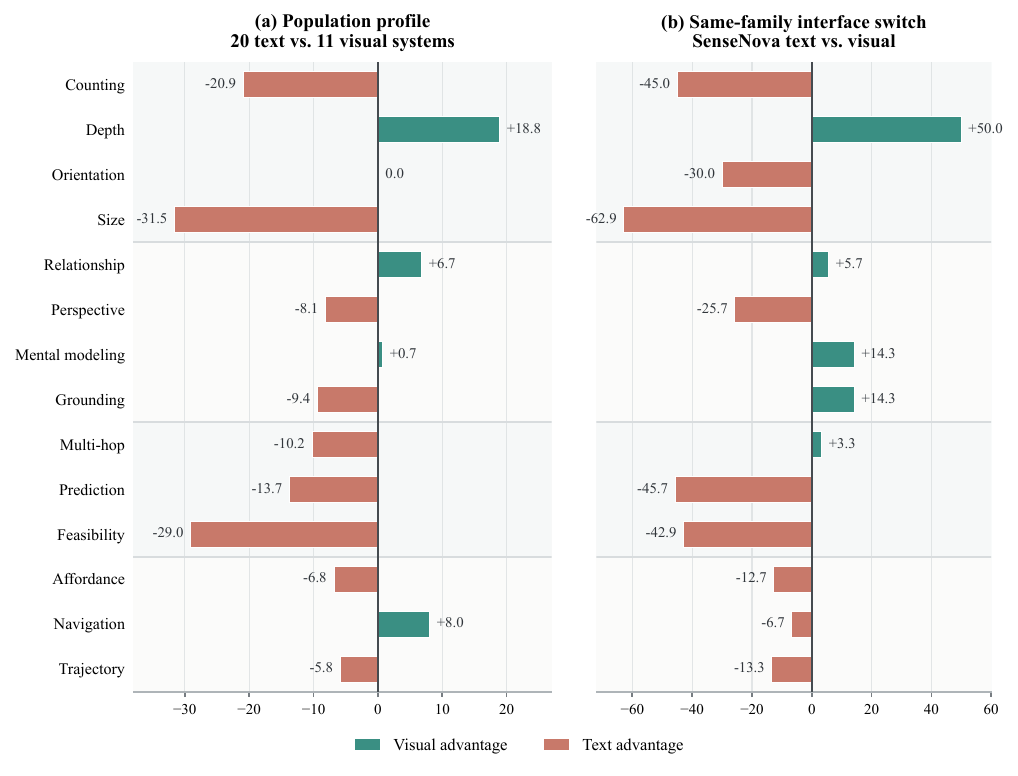}
    \caption{\textbf{Where does showing help?} Bars report visual-minus-text score differences under the same source-task metrics; positive values favor visual answering. (a) compares the text- and visual-model means. (b) provides a partial same-family comparison between the two SenseNova-Vision-7B-MoT interfaces on identical items. Both comparisons are descriptive because their model pools or decoding paths differ.}
    \label{fig:where-showing-helps}
\end{figure}

\textbf{Visual answering is strongest when the target can remain a pixel-level spatial state.}
Both panels of Fig.~\ref{fig:where-showing-helps} show visual gains on Depth (+18.85 in the model-group comparison and +50.00 for SenseNova) and Relationship (+6.74 and +5.71). These tasks allow the answer to remain directly visible: a depth field preserves continuous near--far structure, while grounded overlays keep the referenced entities and their spatial configuration jointly observable. The model can therefore express its conclusion in the same spatial medium as the evidence rather than first reduce it to a symbolic representation. Navigation improves only in the model-group comparison and reverses for SenseNova, so the visual advantage is concentrated in directly externalizable, spatially grounded states rather than visual tasks in general.

\textbf{Text answering remains stronger when solving requires transformation beyond visible evidence.}
Both panels favor text answering on Size, Feasibility, Prediction, and Counting, while the text-model mean leads by 17.63 points at the Reasoning level. These tasks require operations beyond local visual evidence: comparing object extent, testing geometric compatibility, updating a state under spatial dynamics, or aggregating repeated instances. Pixels provide the premises, but the answer emerges only after intermediate relations or constraints are composed into a discrete decision. Current text-output VLMs are therefore stronger at transforming visual observations into inferred spatial conclusions, whereas visual generation is strongest when the desired state can be externalized directly in pixels.

\begin{table}[H]
\centering
\small
\setlength{\tabcolsep}{6.5pt}
\renewcommand{\arraystretch}{1.14}
\caption{Paired sample-level outcomes on the shared benchmark items. Entries report counts and benchmark shares in parentheses. Visual rescue is $\textit{Visual only}/(\textit{Visual only}+\textit{Neither})$: the fraction of text-incorrect items solved by the paired visual interface.}
\label{tab:interface-complementarity}
\begin{tabular*}{0.94\linewidth}{@{\extracolsep{\fill}}l*{5}{c}@{}}
\toprule
\textbf{Pair} & \textbf{Both correct} & \textbf{Text only} & \textbf{Visual only} & \textbf{Neither} & \textbf{Visual rescue} \\
\midrule
GPT-5.4 / GPT Image 2 & 182 (38.7) & 96 (20.4) & \textbf{71 (15.1)} & 121 (25.7) & 37.0\% \\
SenseNova (Text / Visual) & 73 (15.5) & 142 (30.2) & \textbf{78 (16.6)} & 177 (37.7) & 30.6\% \\
\bottomrule
\end{tabular*}

\vspace{0.25em}
\begin{minipage}{0.96\linewidth}
\scriptsize
Correctness follows the declared binary result when available; Affordance uses precision $\geq 0.5$, and Prediction requires the exact endpoint score. This overlap diagnostic does not replace the native leaderboard metrics.
\end{minipage}
\end{table}

\textbf{Visual and textual answering provide complementary capabilities.}
Table~\ref{tab:interface-complementarity} pairs predictions on the same items for the strongest evaluated model under each interface and, separately, for the two SenseNova modes. GPT Image 2 solves 71 items missed by GPT-5.4, recovering 37.0\% of its diagnostic failures; SenseNova's visual mode solves 78 items missed by its text mode, recovering 30.6\% despite a lower aggregate score. Thus, visual successes are not simply a subset of those achieved through text answering. The paired results establish sample-level complementarity, but not a separate visual-only task domain or a causal modality effect. Appendix~\ref{app:qualitative-complementarity} provides representative visual-only cases.

\subsection{Agentic Cross-Benchmark Generalization}
\label{sec:agentic-cross-benchmark}

We test whether the Agentic builder transfers beyond SpatialGen-Bench on EmbSpatial-Bench~\citep{du2024embspatial}, OmniSpatial~\citep{jia2025omnispatial}, Q-Spatial+~\citep{liao2024reasoningpathsreferenceobjects}, RoboSpatial-Home~\citep{song2025robospatial}, SAT~\citep{ray2024sat}, and RoboAfford~\citep{Tang2025RoboAffordAD}. These benchmarks differ in data schema, answer form, task organization, and source metric. For each benchmark, the builder constructs task-specific generation--parser artifacts, after which GPT Image 2 and the text baselines are evaluated on the same task-balanced 50-record pilot. This experiment tests end-to-end protocol transfer rather than full-benchmark model ranking; Appendices~\ref{app:agentic-construction-coverage} and~\ref{app:cross-benchmark-details} report construction coverage and pilot details.

\begin{center}
\begin{minipage}{0.98\linewidth}
\centering
\captionsetup{hypcap=false}
\captionof{table}{
Matched cross-benchmark evaluation after Agentic protocol construction (\%). Each benchmark contributes a task-balanced 50-record subset after answer-contract audit, and all models are evaluated on the exact same record IDs. These pilots test end-to-end transfer rather than full-benchmark ranking. Avg. is the unweighted benchmark mean; bold marks the best score in each column.
}
\label{tab:cross-benchmark-results}
\small
\setlength{\tabcolsep}{3pt}
\renewcommand{\arraystretch}{1.12}
\begin{tabular*}{\linewidth}{@{\extracolsep{\fill}}lc*{7}{c}@{}}
\toprule
\textbf{Model} & \textbf{Output} & \textbf{EmbSpatial} & \textbf{OmniSpatial} & \textbf{Q-Spatial} & \textbf{RoboSpatial} & \textbf{SAT} & \textbf{RoboAfford} & \textbf{Avg.} \\
\midrule
GPT Image 2 & Visual & 42.00 & 44.00 & 68.00 & 46.00 & \textbf{74.00} & \textbf{64.00} & 56.33 \\
GPT-5.4 & Text & 82.00 & 38.00 & 72.00 & \textbf{58.00} & 66.00 & 48.00 & \textbf{60.67} \\
Qwen3-VL-8B & Text & \textbf{84.00} & \textbf{46.00} & \textbf{74.00} & 46.00 & 50.00 & 50.00 & 58.33 \\
InternVL3.5-8B & Text & 78.00 & 44.00 & 56.00 & 56.00 & 42.00 & 36.00 & 52.00 \\
\bottomrule
\end{tabular*}
\end{minipage}
\end{center}

\textbf{The Agentic builder transfers beyond SpatialGen-Bench.}
The constructed protocols support end-to-end, metric-compatible GPT Image 2 evaluation on all six benchmarks and matched comparison with the text baselines. The results broadly echo the main benchmark: text models lead four benchmark columns, whereas GPT Image 2 leads SAT and RoboAfford, so neither answer interface is uniformly superior. This pattern is supporting context; the primary result is successful adaptation across heterogeneous task contracts. Because each column is a task-balanced pilot, the values are not full-benchmark leaderboard estimates.

\subsection{Universal VLM Parser Sensitivity}
\label{sec:vlm-judge-analysis}

To measure parser sensitivity, we apply Qwen3-VL-8B~\citep{bai2025qwen3vltechnicalreport}, Qwen2.5-VL-72B~\citep{bai2025qwen25vltechnicalreport}, and Llama 4 Scout as universal parsers to the same saved outputs from six image models, while holding benchmark targets and source-task metrics fixed. ProVisE uses each task's declared parser; the universal parser receives the generated response and task-specific decoding request, but neither the ground-truth answer nor the ProVisE prediction. Every parser column uses the same records, with unavailable artifacts and invalid parses retained as zeros. Table~\ref{tab:vlm-judge-main} reports the resulting scores and rankings; Appendix~\ref{app:vlm-judge-details} gives input exceptions, matched coverage, and task-level results.

\begin{table}[H]
\centering
\small
\setlength{\tabcolsep}{5.5pt}
\renewcommand{\arraystretch}{1.12}
\caption{\textbf{Universal VLM parser sensitivity.} Every parser column scores the same fixed records for each image model. ProVisE uses each declared task parser; the other columns substitute one universal VLM parser. The 36 unavailable generated artifacts and all invalid parses remain in the denominator as zeros in every column. Scores macro-average tasks; superscripts denote within-column ranks.}
\label{tab:vlm-judge-main}
\begin{tabular*}{0.94\textwidth}{@{\extracolsep{\fill}}l*{4}{c}@{}}
\toprule[0.85pt]
\textbf{Image model} & \textbf{ProVisE} & \textbf{Qwen3-VL-8B} & \textbf{Qwen2.5-VL-72B} & \textbf{Llama 4 Scout} \\
\midrule[0.55pt]
GPT Image 2 & \cellcolor{blue!8}\textbf{51.63}\rank{1} & \cellcolor{blue!8}\textbf{45.67}\rank{1} & 44.95\rank{2} & 40.31\rank{3} \\
Nano Banana 2 & 48.63\rank{2} & 43.40\rank{3} & 44.59\rank{3} & 40.60\rank{2} \\
GPT-5 Image Mini & 40.00\rank{3} & 40.60\rank{5} & 39.64\rank{4} & 37.24\rank{4} \\
Seedream 4.5 & 37.89\rank{4} & 42.86\rank{4} & 37.06\rank{6} & \cellcolor{blue!8}\textbf{41.15}\rank{1} \\
JoyAI-Image & 35.28\rank{5} & 44.22\rank{2} & \cellcolor{blue!8}\textbf{46.90}\rank{1} & 35.67\rank{5} \\
FLUX.2 [klein] 4B & 28.84\rank{6} & 38.34\rank{6} & 39.07\rank{5} & 34.63\rank{6} \\
\bottomrule[0.85pt]
\end{tabular*}
\end{table}

\textbf{Parser substitution changes both scores and rankings.}
ProVisE and Qwen3-VL-8B rank GPT Image 2 first, whereas Qwen2.5-VL-72B ranks JoyAI-Image first and Llama 4 Scout ranks Seedream 4.5 first. Relative to ProVisE, the three universal parsers yield rank correlations of 0.60, 0.31, and 0.60 and mean absolute score changes of 5.87, 5.63, and 5.26 points. Their conditional valid-parse rates remain high (99.54\%, 98.49\%, and 91.96\%), indicating that the shifts arise from semantic interpretation as well as response-format failures. Because parser substitution raises some model scores while lowering others, it is not a uniform calibration offset. The experiment therefore shows that a universal black-box parser adds a model-dependent interpretation layer; it does not rank the universal VLMs or image-generation systems themselves.

\FloatBarrier
\subsection{Failure Attribution: Why Does Showing Fail?}
\label{sec:failure-analysis}

To distinguish spatial-answer errors from failures of visual communication, we assign every response to one of five mutually exclusive outcomes: generation failure, protocol noncompliance, parser failure, incorrect prediction, or correct prediction. Correct and incorrect predictions are distinguished only after the parser recovers a valid answer; this diagnostic does not replace the native task metrics used in the main results. The labels identify the observable failure stage rather than the model's latent reasoning, since noncompliance and parser failure may each have multiple causes. Appendix~\ref{app:failure-attribution-details} gives the precedence rules and treatment of historical records without explicit execution status.

\begin{figure}[t]
    \centering
    \includegraphics[width=0.84\textwidth]{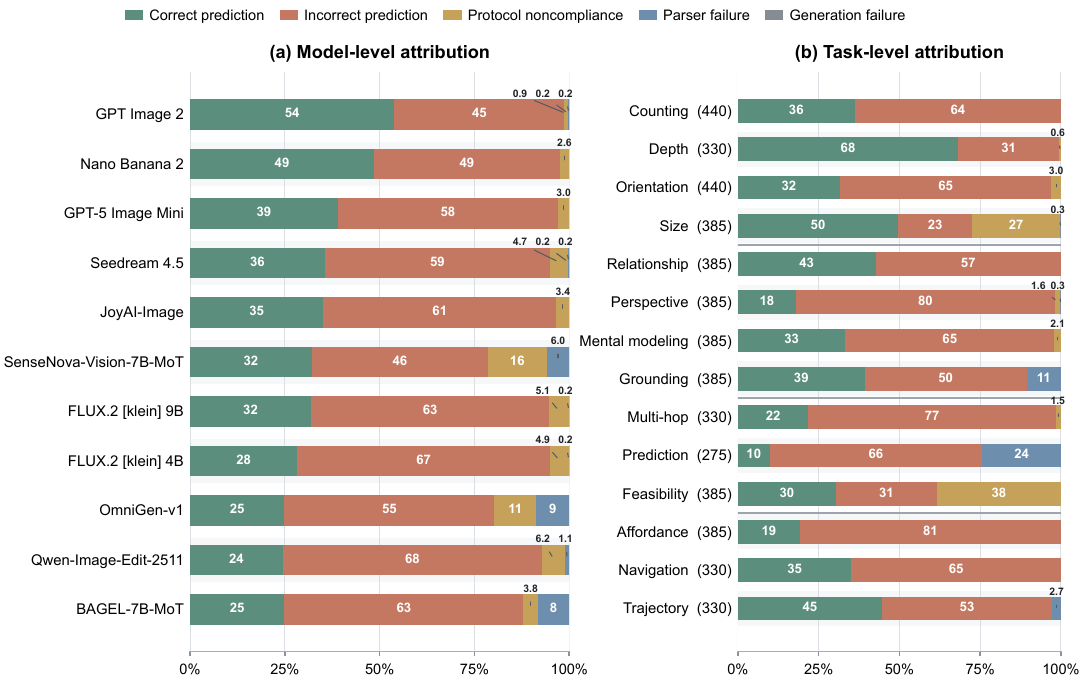}
    \captionsetup{font=footnotesize,skip=4pt}
    \caption{\textbf{Failure attribution by visual-answering model and task.} Bars partition all responses, with correct predictions retained in the denominator. The categories are diagnostic only and do not replace the native metrics in Table~\ref{tab:main-results-vertical}; task separators denote capability groups.}
    \label{fig:failure}
\end{figure}

\textbf{Most failures are valid but incorrect spatial predictions.}
Among non-correct outcomes in Fig.~\ref{fig:failure}, 88.03\% yield valid predictions that fail the diagnostic correctness criterion, compared with 8.46\% protocol noncompliance, 3.45\% parser failure, and 0.06\% generation failure. The visual communication channel therefore usually completes: an image is produced and a task-compatible answer is recovered, but the expressed spatial state is wrong. This distinction explains why ``showing'' is not automatically sufficient. Pixel-space output can preserve a spatial conclusion once formed, but it does not replace the comparison, constraint composition, viewpoint transformation, or state update required to derive that conclusion. The visual interface changes how an answer is expressed, not the reasoning that must precede it.

\textbf{The remaining failures reveal distinct externalization and measurement bottlenecks.}
Non-answer failures are highly localized by task. For Size and Feasibility, 54.64\% and 55.22\% of failures occur before a valid prediction is recovered: the model must preserve the scene while rendering comparative scale or a plausible placement with visible clearance or collision. Prediction and Grounding expose a different burden, with 27.02\% and 17.52\% of failures occurring during answer recovery because endpoints and locations must be encoded precisely enough for parsing. Depth provides the contrasting case: benchmark coordinates supply deterministic measurement anchors, and its parser incurs no parser failures. Showing is therefore most reliable when the answer naturally exists as an inspectable image state and both externalization and measurement are lightweight. It becomes fragile when solving requires hidden transformations, geometrically controlled counterfactuals, or precise visual symbols for downstream recovery.

\FloatBarrier
\begingroup
\titlespacing*{\subsection}{0pt}{1.4ex plus 0.4ex minus 0.2ex}{0.7ex}
\titlespacing*{\section}{0pt}{1.6ex plus 0.4ex minus 0.2ex}{0.8ex}
\setlength{\parskip}{2pt}
\subsection{Limitations}

Agentic protocol construction depends on its planning and validation backends. We use GPT-5.4 and GPT Image 2 as the strongest available backends in our setting to construct one shared protocol for all generators. This improves construction quality and comparability, but may favor visual representations that GPT Image 2 can execute and disadvantage weaker or differently trained generators. The VLM-based Fallback route preserves evaluation coverage when existing recipes or deterministic operators cannot express a task, but its frequent use on some external tasks reveals incomplete library coverage and introduces provider-dependent variability. Expanding reusable recipes, operators, and validation tests is therefore an important direction for the open-source framework. The text- and visual-answering pools also differ in architecture, scale, and training data, so their comparison characterizes current systems rather than isolating a causal modality effect. Finally, ProVisE currently targets static-image benchmarks with recoverable annotations and metrics; extending it to video, simulation, and closed-loop embodied tasks remains future work.

\section{Conclusion}

We introduced ProVisE and SpatialGen-Bench to evaluate spatial cognition beyond text-only answers. ProVisE converts protocol-guided visual responses into structured predictions scored by the original task metrics, while its Agentic builder adapts this process to new benchmarks. SpatialGen-Bench enables a shared comparison of text- and visual-answering models across diverse spatial tasks.

Our results show a division of strengths rather than a single winner. Image-generation models are most competitive when the answer can be represented directly in pixels, such as a depth field or grounded location. Generation then preserves continuous geometry and makes the evidence easy to inspect. Text-answering models remain stronger when visible evidence must be transformed through constraint composition, counterfactual reasoning, or precise state updates. Most failed visual responses are still parseable but encode the wrong answer, showing that visual expression does not replace reasoning. The external pilots demonstrate protocol transfer, while the parser-sensitivity experiment shows that scores and rankings depend on how generated evidence is decoded.

Together, these findings suggest that spatial systems should not force every task into either text or pixels. Text or latent states can handle constraints and planning, visual generation can externalize geometry and grounded state, and deterministic operators can verify the result. Selecting the representation by task and reasoning stage would let ``show'' and ``tell'' exchange intermediate states and cross-check one another.
\endgroup

\bibliographystyle{assets/plainnat}
\bibliography{refs}

\clearpage
\preto\section{\FloatBarrier}
\appendix

\section*{Appendix Overview}

The appendix provides supporting context, implementation details, and complete results:

\begin{itemize}[leftmargin=*]
    \item \textbf{Appendix~\ref{app:related_work}} reviews spatial and generative-model evaluation.
    \item \textbf{Appendix~\ref{app:dataset_construction}} documents data sources, schema, and quality control.
    \item \textbf{Appendix~\ref{app:prompt_parser_protocols}} specifies parser rules, Agentic construction, and validation analyses.
    \item \textbf{Appendix~\ref{app:qualitative-complementarity}} presents representative visual-only successes.
    \item \textbf{Appendix~\ref{app:evaluation_settings}} reports runtime settings, scoring, complete results, and uncertainty.
\end{itemize}

\section{Related Work}
\label{app:related_work}

\noindent\textbf{Spatial reasoning benchmarks and models.}
Spatial evaluation has progressed from single-image relation and perception tests, including VSR, BLINK, and ViewSpatial~\citep{liu2023visual,fu2024blink,li2025viewspatial}, to broader taxonomies such as SpatialBench, OmniSpatial, and SpatialScore~\citep{xu2026spatialbenchbenchmarkingmultimodallarge,jia2025omnispatial,wu2025spatialscore}. Planning and embodied suites probe action-oriented reasoning~\citep{ray2024sat,du2024embspatial,Wu_2025_ICCV}, while VSI-Bench, MindCube, and MMSI extend evaluation to video, sparse views, and multiple images~\citep{yang2025thinking,wang2026mindcubespatialmentalmodeling,yang2025mmsi}. In parallel, spatially specialized models introduce geometric supervision, grounded reasoning, or spatially focused post-training~\citep{chen2024spatialvlm,cheng2024spatialrgpt,wu2025spatial,ouyang2025spacer}. These advances broaden spatial coverage, but their heterogeneous answer schemas and metrics remain oriented primarily toward textual or coordinate outputs.

\noindent\textbf{Generative and unified visual systems.}
Unified models such as Janus, BAGEL, Emu3.5, and Show-o2 combine visual understanding and generation within one system~\citep{chen2025janus,deng2025emerging,cui2025emu3,xie2025show}. Existing generative evaluations measure prompt faithfulness, object composition, or scene-graph consistency through question answering, detectors, or learned judges~\citep{hu2023tifa,ghosh2023geneval,huang2025t2i,cho2024davidsonianscenegraphimproving}. They are not designed to treat a generated image as the answer to a source benchmark whose native metric expects a choice, coordinate, mask, or trajectory. Our setting instead studies metric-compatible visual answering under those heterogeneous task contracts.

\section{Dataset Construction and Quality Control}
\label{app:dataset_construction}

This section documents the construction and quality assessment of the compact SpatialGen-Bench split. The main text already reports the benchmark scale, capability hierarchy, and task coverage in the benchmark construction section and Table~\ref{tab:benchmark_comparison}. We therefore focus here on source provenance, task mapping, unified data representation, and quality control of the final released split.

\subsection{Data Sources, Task Mapping, and Licenses}
\label{app:data_sources_task_mapping}

SpatialGen-Bench is assembled from public spatial benchmarks that cover object-level perception, scene-level spatial understanding, multi-step reasoning, and action-oriented interaction. Each source task is mapped according to its task objective rather than the source benchmark name alone. For example, object-facing direction is assigned to perception because it is directly observable from the image, whereas viewpoint selection is assigned to understanding because it requires reasoning from another perspective. 
We do not repeat the hierarchy figure here; instead, Table~\ref{tab:app_source_license} records the released sample count, source benchmark, original task type, capability mapping, and license or usage terms for every subtask.

All samples retain provenance in \texttt{source}, \texttt{source\_id}, and \texttt{metadata\_json}. We report the usage terms available from the corresponding official project page, repository, or Hugging Face dataset card. Because repository or dataset-card licenses may not cover every upstream media asset, this table documents provenance rather than reinterpreting source rights. ``N/A'' means that no explicit license field was identified in the checked official source; it does not imply absence of restrictions.

\begin{center}
\begin{minipage}{0.98\linewidth}
\centering
\captionsetup{hypcap=false}
\scriptsize
\setlength{\tabcolsep}{2.2pt}
\renewcommand{\arraystretch}{1.10}
\captionof{table}{Source provenance, released sample count ($N$), task mapping, and usage terms for SpatialGen-Bench. Source links point to the checked official project, repository, or dataset card.}
\label{tab:app_source_license}
\begin{tabularx}{\linewidth}{@{}L{0.105\linewidth} C{0.04\linewidth} L{0.145\linewidth} L{0.30\linewidth} C{0.115\linewidth} C{0.09\linewidth} Y@{}}
\toprule
\textbf{Task} & \textbf{$N$} & \textbf{Source benchmark} & \textbf{Original task type} & \textbf{Capability} & \textbf{Terms} & \textbf{Source} \\
\midrule
Counting &
40 &
CountBench &
Object/category counting in a single image. &
Perception &
CC-BY-4.0 &
\href{https://teaching-clip-to-count.github.io/}{Project} \\

Depth &
30 &
BLINK &
A/B relative-depth comparison. &
Perception &
Apache-2.0 &
\href{https://huggingface.co/datasets/BLINK-Benchmark/BLINK}{HF} \\

Orientation &
40 &
EgoOrientBench &
Egocentric object-facing direction. &
Perception &
MIT &
\href{https://huggingface.co/datasets/jhCOR/EgoOrientBench}{HF} \\

Object Size &
35 &
SPHERE-VLM~\citep{zhang2025sphere} &
Object-size comparison in natural images. &
Perception &
COCO terms &
\href{https://huggingface.co/datasets/wei2912/SPHERE-VLM}{HF} \\

\addlinespace[0.25em]
Relationship &
35 &
VSR &
True/false visual-spatial relation verification. &
Understanding &
Apache-2.0 &
\href{https://github.com/cambridgeltl/visual-spatial-reasoning}{GitHub} \\

Perspective &
35 &
ViewSpatial-Bench &
Perspective or viewpoint-direction choice. &
Understanding &
Apache-2.0 &
\href{https://huggingface.co/datasets/lidingm/ViewSpatial-Bench}{HF} \\

Mental Modeling &
35 &
MindCube &
Multi-view mental modeling and perspective reasoning. &
Understanding &
MIT &
\href{https://huggingface.co/datasets/MLL-Lab/MindCube}{HF} \\

Spatial Grounding &
35 &
RefCOCOg~\citep{mao2016generation} &
Referring-expression localization with region annotations. &
Understanding &
N/A &
\href{https://huggingface.co/datasets/lmms-lab/RefCOCOg}{HF} \\

\addlinespace[0.25em]
Multi-hop &
30 &
VisWorld-Eval &
Multi-hop spatial manipulation and compositional reasoning. &
Reasoning &
N/A &
\href{https://huggingface.co/datasets/thuml/VisWorld-Eval}{HF} \\

Prediction &
25 &
VisWorld-Eval &
Synthetic state prediction under spatial dynamics. &
Reasoning &
N/A &
\href{https://huggingface.co/datasets/thuml/VisWorld-Eval}{HF} \\

Geometric Feasibility &
35 &
SPHERE-VLM~\citep{zhang2025sphere} &
Object manipulation, geometric fit, and placement feasibility. &
Reasoning &
COCO terms &
\href{https://huggingface.co/datasets/wei2912/SPHERE-VLM}{HF} \\

\addlinespace[0.25em]
Affordance &
35 &
RoboAfford-Eval &
Affordance or spatial-region grounding. &
Interaction &
CC-BY-4.0 &
\href{https://huggingface.co/datasets/tyb197/RoboAfford-Eval}{HF} \\

Navigation &
30 &
PhysBench &
Candidate future-view selection under motion. &
Interaction &
Apache-2.0 &
\href{https://huggingface.co/datasets/USC-PSI-Lab/PhysBench}{HF} \\

Trajectory &
30 &
ShareRobot-Bench &
Robot manipulation trajectory planning. &
Interaction &
Apache-2.0 &
\href{https://huggingface.co/datasets/BAAI/ShareRobot-Bench}{HF} \\
\midrule
\textbf{Total} & \textbf{470} & \multicolumn{5}{l}{\textit{14 tasks across four capability levels}} \\
\bottomrule
\end{tabularx}
\end{minipage}
\end{center}

\subsection{Unified Instance Format}
\label{app:unified_instance_format}

Each selected item is serialized as one row in the released JSONL. We preserve the source answer semantics while normalizing shared field names, relative media paths, option identifiers, and surface labels required for unified loading and scoring. A depth item therefore remains an A/B comparison, a relation item remains boolean verification, and a trajectory item remains an ordered-path target. The schema standardizes transport; it does not redefine the task contract.

\begin{center}
\begin{minipage}{0.98\linewidth}
\centering
\captionsetup{hypcap=false}
\scriptsize
\setlength{\tabcolsep}{3.5pt}
\renewcommand{\arraystretch}{1.08}
\captionof{table}{Released SpatialGen-Bench JSONL schema. Paths are relative to the dataset root. \texttt{mask\_path} and evaluator metadata are scoring targets, not model inputs.}
\label{tab:app_unified_schema}
\begin{tabularx}{\linewidth}{@{}C{0.20\linewidth} C{0.16\linewidth} X@{}}
\toprule
\textbf{Field} & \textbf{Type} & \textbf{Meaning} \\
\midrule
\texttt{id} & string & Globally unique sample identifier. \\
\texttt{capability} & string & One of perception, understanding, reasoning, or interaction. \\
\texttt{task} & string & One of the 14 normalized subtask names. \\
\texttt{image\_path} & string & Relative path to the primary input image. \\
\texttt{media\_paths} & list[string] & Ordered model-input media, including any required views or candidates. \\
\texttt{mask\_path} & string & Optional evaluator-only target-mask path; empty when unused. \\
\texttt{question} & string & Normalized instruction shown to the model. \\
\texttt{answer} & string & Canonical serialized answer in the source task's answer space. \\
\texttt{answer\_type} & string & Declared answer schema, such as integer, choice, boolean, point, mask, or path. \\
\texttt{choices} & list[string] & Ordered candidates; empty when the task is not choice-based. \\
\texttt{source} & string & Source benchmark identifier. \\
\texttt{source\_id} & string & Identifier retained from the source benchmark. \\
\texttt{metadata\_json} & string & JSON-encoded provenance and task-specific auxiliary metadata. \\
\bottomrule
\end{tabularx}
\end{minipage}
\end{center}

Table~\ref{tab:app_unified_schema} matches the public release rather than the richer in-memory Python object. \texttt{media\_paths} explicitly records ordered model inputs, while \texttt{mask\_path} remains evaluator-only. Additional views, candidate states, object references, start points, paths, source annotations, and curation metadata are retained in \texttt{metadata\_json}; loaders parse that string as JSON. This flat representation remains portable across JSONL and table-based hosting formats. Listing~\ref{ex:sage_jsonl_record} reproduces a released counting record.

\begin{minipage}{\linewidth}
\begin{lstlisting}[
caption={Representative SpatialGen-Bench JSONL record.},
label={ex:sage_jsonl_record}
]
{
  "id": "perception_counting_110",
  "capability": "perception",
  "task": "counting",
  "image_path": "perception/counting/110.jpg",
  "media_paths": ["perception/counting/110.jpg"],
  "mask_path": "",
  "question": "How many dice are there in the image?",
  "answer": "10",
  "answer_type": "integer",
  "choices": [],
  "source": "CountBench",
  "source_id": "perception_counting_110",
  "metadata_json": "{\"text\":\"d10 set of ten green dice\",\"source\":\"CountBench\"}"
}
\end{lstlisting}
\end{minipage}

\subsection{Data Integrity and Review Coverage}
\label{app:source_data_quality_assessment}

Deterministic integrity checks cover the entire released split and verify unique IDs, relative paths, readable primary media, answer-schema validity, placeholder-like assets, and required task-specific inputs or evaluator annotations; every retained record passes. Manual inclusion review likewise covers every item and checks that the source instruction, visual evidence, annotation, and scoring contract remain mutually consistent.

\section{Protocol Design and Parsing}
\label{app:prompt_parser_protocols}

This section specifies the protocol constraints, parser rules, Agentic routes, and validation evidence used in evaluation. Exact prompts remain executable artifacts in the released code rather than a duplicated prompt catalog.

\subsection{Protocol Design}
\label{app:visual_prompt_templates}

Per-sample prompts substitute only declared benchmark fields such as \texttt{question}, \texttt{choices}, object references, candidate states, or trajectory start points. Newly constructed Agentic protocols follow four invariants:
\begin{itemize}[leftmargin=1.5em,itemsep=0.2em,topsep=0.35em]
    \item \textbf{Task-grounded evidence.} The response must expose a location, region, relation, state, measurement, or path that is meaningful for the source task.
    \item \textbf{Minimal intervention.} Editing protocols preserve unrelated scene content; synthesis protocols retain the task's scene and viewpoint constraints.
    \item \textbf{Executable parser.} Every prompt is paired with a declared deterministic, fixed-auxiliary, or generated-image-only parser and explicit invalid-output conditions.
    \item \textbf{No answer shortcut.} The builder rejects option letters, corner codes, verdict symbols, or prose-only outputs when these merely encode the answer without spatial evidence.
\end{itemize}

SpatialGen-Bench retains fixed-slot protocols for Mental Modeling and Multi-hop, and a trajectory-guided endpoint protocol for Prediction. In the latter, the generated response draws one reflected yellow path and circles one destination hole; a fixed auxiliary VLM then recovers the expressed hole index and candidate count from that image alone. These evaluated protocols are distinct from the Agentic routes constructed for the external benchmarks.

Figures~\ref{fig:protocolized_workflow} and~\ref{fig:protocol_examples} summarize the protocol pool and representative outputs. Full prompt specifications and benchmark-specific YAML artifacts are available in the project repository linked on the first page.

\subsection{Parser Rules and Thresholds}
\label{app:parser_rules}

Table~\ref{tab:app_parser_rules} separates reusable parsers from those instantiated or selected during Agentic construction. It records each operational rule, decoded answer type, and invalid-output boundary; CLIP, OCR, and VLM assistance are named explicitly rather than grouped under a generic model-assisted parser label.

\begin{center}
\begin{minipage}{0.98\linewidth}
\centering
\captionsetup{hypcap=false}
\scriptsize
\setlength{\tabcolsep}{2.8pt}
\renewcommand{\arraystretch}{1.04}
\captionof{table}{Parser rules for reusable and Agentic protocols. HSV ranges use OpenCV hue values; fixed auxiliary models and generated-image-only Fallback are named explicitly.}
\label{tab:app_parser_rules}
\begin{tabularx}{\linewidth}{@{}C{0.14\linewidth} L{0.39\linewidth} C{0.16\linewidth} >{\raggedright\arraybackslash}X@{}}
\toprule
\textbf{Protocol} & \textbf{Parser rule} & \textbf{Prediction} & \textbf{Invalid or unreliable condition} \\
\midrule
\rowcolor{blue!5}
\multicolumn{4}{@{}l}{\textbf{Reusable SpatialGen-Bench protocols}} \\
Instance Marking & Threshold green pixels (H=35--90, S/V$\geq$80), apply $3\times3$ morphology, and retain plausible connected components by area and solidity. & Integer count. & No valid component, merged markers, or marks not corresponding to target instances. \\
Direction Grid & Threshold blue pixels (H=100--130, S$\geq$120, V$\geq$50), aggregate evidence over the eight outer cells of a $3\times3$ grid, and require at least 50 blue pixels. & Direction label. & No active cell, center-cell marking, or conflicting evidence across cells. \\
Fixed-Slot Coding & Threshold magenta pixels (H=140--175, S/V$\geq$80) in the protocol's predefined slots and require at least 30 pixels. & Choice label or numeric slot. & Missing mark, multiple active slots, or a mark outside the code layout. \\
Binary Relation & Recover green subject and blue object overlays using fixed HSV ranges and minimum color support. & Boolean label. & Wrong entities are marked, required grounded masks are absent, or scene colors create an unsupported signal. \\
Relative Depth & Sample $5\times5$ local grayscale means at the benchmark A/B locations and select the brighter location as closer. & A/B depth label. & Unreadable depth map, invalid locations, indistinguishable local values, or violation of the declared brightness convention. \\
Region Masking & Resize to the target resolution, threshold grayscale at $>127$, and pass the binary mask to the source metric. & Binary mask. & Empty or unreadable mask, missing target metadata, or non-binary artifacts. \\
State Matching & Encode the generated state and candidate states with fixed CLIP ViT-B/32 and select the maximum cosine similarity. & Candidate label. & Missing candidates, unreadable images, or a generated state not comparable to the candidates. \\
Trajectory Drawing & Extract the novel red component nearest the start point, skeletonize and order it, then sample 20 normalized path points. & Normalized polyline. & Missing or disconnected path, wrong start, or excessive red artifacts. \\
Trajectory-Guided Endpoint & A fixed auxiliary VLM reads the numbered hole enclosed by the blue circle and the total hole count from the generated yellow-trajectory response alone. & Indexed endpoint. & Missing or ambiguous endpoint circle, unreadable hole numbering, or a non-positive endpoint/count. \\
Point Marker & Threshold novel cyan pixels (H=80--100, S/V$\geq$70), select one compact component, and normalize its centroid. & Normalized point. & Missing marker, multiple similarly prominent marks, or a component too small or diffuse to localize. \\
\addlinespace[0.3em]
\rowcolor{green!5}
\multicolumn{4}{@{}l}{\textbf{Agentic Build and Fallback parsers}} \\
Marked-Object Choice & Align the output to the source, subtract pre-existing magenta, recover one contour, crop the enclosed source object, and match it to textual choices with fixed CLIP. & Choice label. & Missing or ambiguous contour, invalid crop, or insufficient CLIP separation. \\
Grounded Dimension & Recover magenta/cyan anchors and the yellow dimension line by color differencing; local OCR parses the adjacent scalar-unit tag. & Scalar and unit. & Required anchors or the dimension line are missing; OCR fails; the unit is unsupported; or the label is detached. \\
Dual-Anchor Relation & Recover role-colored subject and reference masks, then compute the configured horizontal, vertical, dominant-axis, or distance relation. & Choice label. & Either anchor is missing or ambiguous, or the recovered geometry cannot map uniquely to a choice. \\
Relation-Zone Boolean & Recover subject, reference, and valid-relation-region masks; predict true exactly when the subject lies in the rendered valid region. & Boolean label. & Missing anchors, degenerate relation region, or a verdict-only symbol instead of spatial evidence. \\
Size Fallback & A constrained parser maps only visible object-comparison or measurement evidence in the generated response to the A/B schema. & A/B size label. & Bare label, missing comparison evidence, contradictory marks, or ambiguous referenced objects. \\
Feasibility Fallback & A constrained parser reads only the depicted placement, clearance, overlap, or collision state. & Boolean label. & Attempted placement absent, unrealistic or ambiguous scale, verdict symbol, or no visible fit/collision evidence. \\
Generic Fallback & A task-specific parser prompt recovers the declared answer schema solely from the generated response; choices may only map visible evidence to labels. & Contract-specific answer. & Evidence absent or contradictory, output contains only prose/label, or reading would require the source image or original question. \\
\bottomrule
\end{tabularx}
\end{minipage}
\end{center}

\subsection{Agentic Construction Details and Coverage}
\label{app:agentic-construction-coverage}

\begin{center}
\begin{minipage}{0.98\linewidth}
\centering
\captionsetup{hypcap=false}
\footnotesize
\setlength{\tabcolsep}{3pt}
\renewcommand{\arraystretch}{1.08}
\captionof{table}{
Agentic protocol construction on six external spatial benchmarks after answer-contract audit. Records, task units, routes, and Smoke describe the current protocol artifacts; Pilot and Valid summarize the archived 50-record-per-benchmark GPT Image 2 evaluation. Invalid visual outputs remain in the denominator and receive zero score. Smoke reports protocol executability, not target-model accuracy.
}
\label{tab:agentic-adaptation}
\begin{tabular*}{\linewidth}{@{\extracolsep{\fill}}l*{8}{c}@{}}
\toprule
\textbf{Benchmark} & \textbf{Records} & \textbf{Tasks} & \textbf{Reuse} & \textbf{Build} & \textbf{Fallback} & \textbf{Pilot} & \textbf{Valid (\%)} & \textbf{Smoke} \\
\midrule
EmbSpatial-Bench & 3,640 & 1 & 0 & 1 & 0 & 50 & 100.00 & 1/1 \\
OmniSpatial & 1,533 & 5 & 0 & 0 & 5 & 50 & 82.00 & 5/5 \\
Q-Spatial+ & 101 & 3 & 0 & 3 & 0 & 50 & 100.00 & 3/3 \\
RoboSpatial-Home & 350 & 3 & 1 & 2 & 0 & 50 & 100.00 & 3/3 \\
SAT & 150 & 8 & 0 & 5 & 3 & 50 & 92.00 & 8/8 \\
RoboAfford & 338 & 3 & 3 & 0 & 0 & 50 & 96.00 & 3/3 \\
\midrule
\textbf{Total} & \textbf{6,112} & \textbf{23} & \textbf{4} & \textbf{11} & \textbf{8} & \textbf{300} & \textbf{95.00} & \textbf{23/23} \\ 
\bottomrule
\end{tabular*}
\end{minipage}
\end{center}

\noindent\textbf{Build contract.}
For each task, the construction Agent receives up to three answer-hidden examples together with the observed input and answer schemas, candidate choices, metric contract, and registered protocol and Parser Ops inventories. It returns a structured configuration that Python compiles into generation guidance, an executable parser, invalid-output rules, and metric mapping. Static checks reject unknown operators, type or metric mismatches, missing media, and answer leakage; ground-truth answers and correctness feedback are withheld from both the initial and revision calls.

The implementation registers 11 reusable high-level protocols and 41 typed Parser Ops covering color, components, differencing, geometry, points, masks, paths, measurements, OCR, choice mapping, and fixed CLIP similarity. The terminal \texttt{unsupported} state is reserved for inputs that cannot be normalized or scored and therefore produces no evaluation result.

\noindent\textbf{Smoke gates.}
Smoke validation generates up to three responses and parses each twice. Acceptance requires successful generation and parsing, repeated-parse agreement, recoverable spatial evidence, and metric-compatible predictions; a Fallback route must also yield at least one valid prediction. Correctness is diagnostic rather than an acceptance gate. Parser or contract failures permit one revision, whereas generation or API failures are deferred. Accepted YAML and manifest artifacts are reused across target models without rerouting.

\noindent\textbf{Coverage.}
Table~\ref{tab:agentic-adaptation} summarizes benchmark-level construction and pilot validity. After answer-contract audit, all 23 task units pass smoke validation: four Reuse routes and 11 Build routes use deterministic or fixed-auxiliary parsers, while eight use generated-image-only Fallback. Table~\ref{tab:agentic-task-routes} reports every task-level decision, construction primitive, parser, metric, and acceptance stage.

\begin{landscape}
\begin{center}
\captionsetup{hypcap=false}
\captionof{table}{Task-level Agentic protocol construction on the six external spatial benchmarks after contract audit. Route identifies whether the Agent selected a validated protocol (Reuse), instantiated a registered parser recipe (Build), or emitted a generated-image-only parser contract (Fallback). Smoke stage reports where each protocol artifact passed validation; it is not an accuracy result.}
\label{tab:agentic-task-routes}
\small
\setlength{\tabcolsep}{3.0pt}
\renewcommand{\arraystretch}{1.30}
\begin{tabularx}{0.98\linewidth}{@{}L{0.14\linewidth} >{\raggedright\arraybackslash}X C{0.045\linewidth} C{0.07\linewidth} L{0.17\linewidth} C{0.13\linewidth} C{0.11\linewidth} C{0.11\linewidth}@{}}
\toprule
\textbf{Benchmark} & \textbf{Task contract} & \textbf{$N$} & \textbf{Route} & \textbf{Construction} & \textbf{Parser} & \textbf{Metric} & \textbf{Smoke stage} \\
\midrule
EmbSpatial-Bench & Spatial relationship identification & 3,640 & Build & Marked-object choice & CV + CLIP & Accuracy & Pass (initial) \\
\addlinespace[0.25em]
OmniSpatial & Dynamic reasoning & 420 & Fallback & Task-specific prompt pair & Generated-only VLM & Accuracy & Pass (initial) \\
 & Perspective taking & 561 & Fallback & Task-specific prompt pair & Generated-only VLM & Accuracy & Pass (transition) \\
 & Spatial interaction & 300 & Fallback & Task-specific prompt pair & Generated-only VLM & Accuracy & Pass (initial) \\
 & Complex logic (choice) & 23 & Fallback & Task-specific prompt pair & Generated-only VLM & Accuracy & Pass (initial) \\
 & Complex logic (structured/text) & 229 & Fallback & Task-specific prompt pair & Generated-only VLM & Accuracy & Pass (initial) \\
\addlinespace[0.25em]
Q-Spatial+ & 1-D horizontal extent & 1 & Build & Grounded dimension & CV + OCR & Ratio success & Pass (initial) \\
 & Horizontal distance & 98 & Build & Grounded dimension & CV + OCR & Ratio success & Pass (initial) \\
 & Vertical distance & 2 & Build & Grounded dimension & CV + OCR & Ratio success & Pass (initial) \\
\addlinespace[0.25em]
RoboSpatial-Home & Compatibility & 105 & Build & Relation-zone boolean & CV geometry & Accuracy & Pass (initial) \\
 & Configuration & 123 & Build & Relation-zone boolean & CV geometry & Accuracy & Pass (initial) \\
 & Context grounding & 122 & Reuse & Point marker & CV centroid & Point-in-mask & Pass (initial) \\
\addlinespace[0.25em]
SAT & Action consequence (binary) & 32 & Build & Relation-zone boolean & CV geometry & Accuracy & Pass (initial) \\
 & Action consequence (choice) & 5 & Fallback & Task-specific prompt pair & Generated-only VLM & Accuracy & Pass (revision) \\
 & Ego movement & 23 & Fallback & Task-specific prompt pair & Generated-only VLM & Accuracy & Pass (revision) \\
 & Goal aim (binary) & 4 & Build & Relation-zone boolean & CV geometry & Accuracy & Pass (initial) \\
 & Goal aim (choice) & 30 & Build & Dual-anchor relation & CV geometry & Accuracy & Pass (initial) \\
 & Object movement & 23 & Fallback & Task-specific prompt pair & Generated-only VLM & Accuracy & Pass (revision) \\
 & Perspective (binary) & 7 & Build & Relation-zone boolean & CV geometry & Accuracy & Pass (initial) \\
 & Perspective (choice) & 26 & Build & Dual-anchor relation & CV geometry & Accuracy & Pass (initial) \\
\addlinespace[0.25em]
RoboAfford & Object affordance & 124 & Reuse & Point marker & CV centroid & Point-in-mask & Pass (initial) \\
 & Object reference & 114 & Reuse & Point marker & CV centroid & Point-in-mask & Pass (initial) \\
 & Spatial affordance & 100 & Reuse & Point marker & CV centroid & Point-in-mask & Pass (initial) \\
\bottomrule
\end{tabularx}
\end{center}
\end{landscape}

\subsection{Cross-Benchmark Evaluation Details}
\label{app:cross-benchmark-details}

Table~\ref{tab:cross-benchmark-results} uses 50 archived GPT Image 2 record IDs per benchmark, selected by deterministic task-balanced allocation, and evaluates each text baseline on the same IDs. Parsing succeeds for 285/300 visual outputs (95.00\%): deterministic or fixed-auxiliary routes recover 225/231 (97.40\%), and Fallback recovers 60/69 (86.96\%). The remaining outputs comprise six parser failures and nine protocol-noncompliant responses, all retained as zeros; these rates measure protocol executability rather than answer correctness.

OmniSpatial is the only scope exception: its archived pilot covers four task units, whereas current construction coverage includes a fifth restored unit. Unmatched full-split text scores are omitted because their sample composition differs from the visual subsets.

\subsection{Depth Parser Route Ablation}
\label{app:depth-parser-ablation}

SpatialGen-Bench provides normalized A/B query locations for every Depth record. They identify where to read the generated depth map without revealing which point is closer. The formal parser samples a $5\times5$ local grayscale mean at each location, applies the prompt's fixed brighter-is-closer convention, and records both measurements without a VLM call.

Table~\ref{tab:depth-parser-ablation} compares benchmark-coordinate grayscale sampling with the prior mixed route over the same saved outputs. The prior route used cached coordinates when available and otherwise asked a VLM to judge A/B directly from the source marker image and generated depth map. Coordinate sampling yields a higher parse rate and benchmark accuracy, supporting its use as the formal Depth parser while keeping the final decision tied to explicit evidence in the generated map.

Two outputs from the FLUX.2 [klein] variants have zero-valued neighborhoods at both locations and are marked invalid because they provide no distinguishable depth evidence. As shown in Table~\ref{tab:depth-kernel-sensitivity}, aggregate accuracy across kernels from $1\times1$ to $11\times11$ spans only 1.21 points; we retain the existing $5\times5$ default rather than select a kernel retrospectively. The 100\% agreement between recorded grayscale evidence and parser predictions is an implementation check, not an independent accuracy result.

\begin{table}[H]
\centering
\footnotesize
\setlength{\tabcolsep}{6.0pt}
\renewcommand{\arraystretch}{1.08}
\caption{Depth parser route comparison. The formal route samples the generated depth map at benchmark-supplied A/B locations. The prior mixed route uses coordinates when available and otherwise asks a VLM to judge the source marker image and generated depth map directly. Both cover the same 330 model--record pairs; invalid parses remain in the denominator as zeros.}
\label{tab:depth-parser-ablation}

\begin{tabular}{@{\hspace{6pt}}lrr@{\hspace{6pt}}}
\toprule
\textbf{Parser route} & \textbf{Parse (\%)} & \textbf{Accuracy (\%)} \\
\midrule
Benchmark-coordinate grayscale sampling ($5\times5$) & 99.39 & 68.18 \\
Prior mixed coordinate/VLM route & 78.48 & 55.45 \\
\bottomrule
\end{tabular}
\end{table}

\begin{table}[H]
\centering
\footnotesize
\setlength{\tabcolsep}{5.5pt}
\renewcommand{\arraystretch}{1.08}
\caption{Depth sampling-kernel sensitivity. Results use the same 330 model--record pairs; $5\times5$ is the protocol default.}
\label{tab:depth-kernel-sensitivity}

\begin{tabular}{@{\hspace{5pt}}lrrrrrr@{\hspace{5pt}}}
\toprule
\textbf{Kernel} & $1\times1$ & $3\times3$ & $\mathbf{5\times5}$ & $7\times7$ & $9\times9$ & $11\times11$ \\
\midrule
Parse (\%) & 98.79 & 99.39 & \textbf{99.39} & 99.39 & 99.39 & 99.39 \\
Accuracy (\%) & 68.79 & 68.18 & \textbf{68.18} & 68.18 & 69.09 & 69.39 \\
\bottomrule
\end{tabular}
\end{table}

\subsection{Universal VLM Parser Sensitivity Details}
\label{app:vlm-judge-details}

Table~\ref{tab:vlm-judge} reports the full task-level comparison for each image model under the matched availability boundary.

Aggregation requires an exact per-task ID match between the benchmark and every parser cache, rejecting duplicate or missing IDs, unresolved API errors, out-of-range scores, and inconsistent generated-image paths. The ProVisE column uses the deterministic Depth results from Appendix~\ref{app:depth-parser-ablation} and reproduces every declared task score to within 0.005 points before applying the shared availability mask.

Depth is the only input exception: the archived universal-parser caches pair the source marker image with the generated depth map and ask for A/B directly. Excluding Depth preserves the sensitivity result (Spearman 0.71/0.43/0.43; mean absolute shifts 6.40/5.59/5.74 points), so the all-task finding is not driven by this route.

\begin{landscape}
\vspace*{\fill}
\begin{center}
\captionsetup{hypcap=false}
\captionof{table}{Task-level universal VLM parser sensitivity. Overall macro-averages task scores. Parse is the fraction of all expected records normalized into the required answer structure; unavailable images and invalid responses remain explicit zero-scored records.}
\label{tab:vlm-judge}
\scriptsize
\setlength{\tabcolsep}{5.3pt}
\renewcommand{\arraystretch}{1.38}
\begin{tabular}{@{\hspace{4pt}}llccc*{14}{c}@{\hspace{4pt}}}
\toprule
\textbf{Image model} & \textbf{Scorer} & \textbf{Overall} & \textbf{Rank} & \textbf{Parse} & \textbf{Cnt.} & \textbf{Dep.} & \textbf{Ori.} & \textbf{Size} & \textbf{Rel.} & \textbf{Pers.} & \textbf{Mental} & \textbf{Ground} & \textbf{M-Hop} & \textbf{Pred.} & \textbf{Feas.} & \textbf{Aff.} & \textbf{Nav.} & \textbf{Traj.} \\
\midrule
GPT Image 2 & ProVisE & 51.63 & 1 & -- & 82.5 & 63.3 & 52.5 & 80.0 & 45.7 & 40.0 & 25.7 & 77.1 & 0.0 & 54.1 & 31.4 & 60.4 & 40.0 & 70.0 \\
 & Qwen3-VL-8B & 45.67 & 1 & 93.2 & 70.0 & 53.3 & 40.0 & 80.0 & 42.9 & 28.6 & 40.0 & 68.6 & 0.0 & 56.3 & 45.7 & 30.7 & 16.7 & 66.7 \\
 & Qwen2.5-VL-72B & 44.95 & 2 & 91.7 & 75.0 & 50.0 & 55.0 & 82.9 & 51.4 & 45.7 & 31.4 & 62.9 & 0.0 & 50.1 & 71.4 & 26.9 & 10.0 & 16.7 \\
 & Llama 4 Scout & 40.31 & 3 & 86.0 & 67.5 & 80.0 & 52.5 & 74.3 & 40.0 & 28.6 & 37.1 & 22.9 & 0.0 & 58.1 & 37.1 & 12.9 & 23.3 & 30.0 \\
\rowcolor{gray!6}Nano Banana 2 & ProVisE & 48.63 & 2 & -- & 82.5 & 86.7 & 62.5 & 60.0 & 45.7 & 25.7 & 31.4 & 65.7 & 16.7 & 21.0 & 37.1 & 42.5 & 40.0 & 63.3 \\
\rowcolor{gray!6} & Qwen3-VL-8B & 43.40 & 3 & 98.5 & 67.5 & 56.7 & 50.0 & 77.1 & 37.1 & 25.7 & 40.0 & 60.0 & 30.0 & 19.2 & 51.4 & 22.9 & 23.3 & 46.7 \\
\rowcolor{gray!6} & Qwen2.5-VL-72B & 44.59 & 3 & 98.1 & 72.5 & 50.0 & 55.0 & 77.1 & 48.6 & 28.6 & 42.9 & 54.3 & 33.3 & 30.1 & 57.1 & 18.1 & 10.0 & 46.7 \\
\rowcolor{gray!6} & Llama 4 Scout & 40.60 & 2 & 91.5 & 67.5 & 76.7 & 50.0 & 60.0 & 40.0 & 42.9 & 37.1 & 37.1 & 16.7 & 39.3 & 37.1 & 13.9 & 26.7 & 23.3 \\
GPT-5 Image Mini & ProVisE & 40.00 & 3 & -- & 52.5 & 73.3 & 77.5 & 54.3 & 42.9 & 28.6 & 28.6 & 34.3 & 10.0 & 30.7 & 40.0 & 14.1 & 36.7 & 36.7 \\
 & Qwen3-VL-8B & 40.60 & 5 & 99.8 & 62.5 & 53.3 & 50.0 & 68.6 & 45.7 & 40.0 & 34.3 & 34.3 & 33.3 & 35.3 & 51.4 & 16.4 & 23.3 & 20.0 \\
 & Qwen2.5-VL-72B & 39.64 & 4 & 98.7 & 55.0 & 63.3 & 52.5 & 62.9 & 48.6 & 28.6 & 31.4 & 34.3 & 36.7 & 35.1 & 68.6 & 14.8 & 10.0 & 13.3 \\
 & Llama 4 Scout & 37.24 & 4 & 92.3 & 52.5 & 66.7 & 77.5 & 57.1 & 40.0 & 28.6 & 28.6 & 22.9 & 20.0 & 42.5 & 37.1 & 14.7 & 20.0 & 13.3 \\
\rowcolor{gray!6}Seedream 4.5 & ProVisE & 37.89 & 4 & -- & 30.0 & 90.0 & 40.0 & 45.7 & 45.7 & 20.0 & 25.7 & 51.4 & 6.7 & 36.1 & 40.0 & 9.1 & 36.7 & 53.3 \\
\rowcolor{gray!6} & Qwen3-VL-8B & 42.86 & 4 & 98.9 & 50.0 & 50.0 & 35.0 & 82.9 & 51.4 & 31.4 & 37.1 & 37.1 & 36.7 & 41.9 & 48.6 & 27.9 & 30.0 & 40.0 \\
\rowcolor{gray!6} & Qwen2.5-VL-72B & 37.06 & 6 & 97.4 & 45.0 & 40.0 & 32.5 & 74.3 & 51.4 & 40.0 & 22.9 & 37.1 & 20.0 & 32.9 & 57.1 & 15.6 & 10.0 & 40.0 \\
\rowcolor{gray!6} & Llama 4 Scout & 41.15 & 1 & 93.0 & 47.5 & 86.7 & 37.5 & 74.3 & 37.1 & 40.0 & 31.4 & 31.4 & 20.0 & 51.5 & 48.6 & 16.7 & 20.0 & 33.3 \\
JoyAI-Image & ProVisE & 35.28 & 5 & -- & 55.0 & 53.3 & 5.0 & 54.3 & 40.0 & 17.1 & 25.7 & 57.1 & 10.0 & 14.2 & 34.3 & 34.5 & 33.3 & 60.0 \\
 & Qwen3-VL-8B & 44.22 & 2 & 99.8 & 62.5 & 60.0 & 45.0 & 74.3 & 60.0 & 37.1 & 31.4 & 57.1 & 30.0 & 20.6 & 45.7 & 31.9 & 20.0 & 43.3 \\
 & Qwen2.5-VL-72B & 46.90 & 1 & 98.9 & 55.0 & 63.3 & 60.0 & 77.1 & 54.3 & 31.4 & 42.9 & 37.1 & 36.7 & 30.7 & 62.9 & 35.2 & 23.3 & 46.7 \\
 & Llama 4 Scout & 35.67 & 5 & 90.9 & 42.5 & 73.3 & 42.5 & 68.6 & 42.9 & 31.4 & 34.3 & 22.9 & 20.0 & 31.8 & 34.3 & 11.5 & 20.0 & 23.3 \\
\rowcolor{gray!6}FLUX.2 [klein] 4B & ProVisE & 28.84 & 6 & -- & 20.0 & 46.7 & 32.5 & 68.6 & 40.0 & 14.3 & 31.4 & 22.9 & 6.7 & 18.5 & 17.1 & 11.8 & 36.7 & 36.7 \\
\rowcolor{gray!6} & Qwen3-VL-8B & 38.34 & 6 & 99.4 & 47.5 & 53.3 & 20.0 & 71.4 & 45.7 & 37.1 & 45.7 & 31.4 & 30.0 & 30.8 & 51.4 & 29.0 & 20.0 & 23.3 \\
\rowcolor{gray!6} & Qwen2.5-VL-72B & 39.07 & 5 & 98.5 & 57.5 & 50.0 & 37.5 & 65.7 & 40.0 & 28.6 & 37.1 & 40.0 & 26.7 & 30.1 & 68.6 & 18.6 & 20.0 & 26.7 \\
\rowcolor{gray!6} & Llama 4 Scout & 34.63 & 6 & 91.1 & 42.5 & 53.3 & 45.0 & 57.1 & 42.9 & 37.1 & 34.3 & 28.6 & 23.3 & 26.6 & 37.1 & 10.3 & 23.3 & 23.3 \\
\bottomrule
\end{tabular}

\vspace{0.3em}
\footnotesize
Cnt.: Counting; Dep.: Depth; Ori.: Orientation; Rel.: Relationship; Pers.: Perspective; Ground: Grounding; M-Hop: Multi-hop; Pred.: Prediction; Feas.: Feasibility; Aff.: Affordance; Nav.: Navigation; Traj.: Trajectory.
\end{center}
\vspace*{\fill}
\end{landscape}

Of the 2,820 expected model--record pairs per parser, 2,784 retain a cached generated artifact and 36 do not: 34 multihop, one perspective, and one size record, concentrated in GPT Image 2 (31) and Seedream 4.5 (5). These records receive zero under all four parser columns.

Qwen3-VL-8B returns 2,771 valid predictions, 13 invalid predictions, and 36 unavailable-image records; the corresponding counts are 2,742/42/36 for Qwen2.5-VL-72B and 2,560/224/36 for Llama 4 Scout. The valid-parse rate in Table~\ref{tab:vlm-judge} uses all expected records as its denominator. Conditional on an available image, the rates are 99.54\%, 98.49\%, and 91.96\%, respectively. Score and rank changes therefore persist beyond response-format failures.

\subsection{Failure Attribution Details}
\label{app:failure-attribution-details}

The analysis in Sec.~\ref{sec:failure-analysis} uses cached outputs only. Labels follow the precedence generation failure, protocol noncompliance, parser failure, and then correct or incorrect prediction. An invalid or ambiguous Fallback result is noncompliance when the parser executes but finds no recoverable evidence; an exception or unsuccessful parse is parser failure. For Depth, an exact local tie is noncompliance because the generated map provides no comparison. Historical blank fixed-auxiliary outputs are assigned conservatively to parser failure, whereas blank deterministic outputs indicate missing protocol evidence.

Explicit binary correctness is retained when available; otherwise, the diagnostic uses precision $\geq 0.5$ for Affordance, DFD $<0.4$ for Trajectory, and normalized score $\geq 0.999$ elsewhere. These thresholds affect attribution only, not native benchmark scores. Thirty-two records require the conservative inference above; missing historical images are not reclassified as generation failures when cached predictions and scores remain available.

Tables~\ref{tab:failure-by-model} and~\ref{tab:failure-by-task} provide the decompositions underlying Fig.~\ref{fig:failure}.

\begin{table}[H]
\centering
\scriptsize
\setlength{\tabcolsep}{3.8pt}
\renewcommand{\arraystretch}{1.08}
\caption{Failure attribution by visual-answering system. Each cell reports count (row percentage); every system is evaluated on the same SpatialGen-Bench items. Correct prediction is a diagnostic category, not the native benchmark score.}
\label{tab:failure-by-model}
\begin{tabular}{@{\hspace{5pt}}l r r r r r r@{\hspace{5pt}}}
\toprule
\textbf{Model} & \textbf{$N$} & \makecell{\textbf{Correct}\\\textbf{prediction}} & \makecell{\textbf{Incorrect}\\\textbf{prediction}} & \makecell{\textbf{Protocol}\\\textbf{noncompliance}} & \makecell{\textbf{Parser}\\\textbf{failure}} & \makecell{\textbf{Generation}\\\textbf{failure}} \\
\midrule
GPT Image 2 & 470 & 253 (53.8) & 211 (44.9) & 4 (0.9) & 1 (0.2) & 1 (0.2) \\
Nano Banana 2 & 470 & 228 (48.5) & 230 (48.9) & 12 (2.6) & 0 (0.0) & 0 (0.0) \\
GPT-5 Image Mini & 470 & 183 (38.9) & 273 (58.1) & 14 (3.0) & 0 (0.0) & 0 (0.0) \\
Seedream 4.5 & 470 & 167 (35.5) & 279 (59.4) & 22 (4.7) & 1 (0.2) & 1 (0.2) \\
JoyAI-Image & 470 & 165 (35.1) & 289 (61.5) & 16 (3.4) & 0 (0.0) & 0 (0.0) \\
SenseNova-Vision-7B-MoT & 470 & 151 (32.1) & 218 (46.4) & 73 (15.5) & 28 (6.0) & 0 (0.0) \\
FLUX.2 [klein] 9B & 470 & 150 (31.9) & 295 (62.8) & 24 (5.1) & 1 (0.2) & 0 (0.0) \\
FLUX.2 [klein] 4B & 470 & 133 (28.3) & 313 (66.6) & 23 (4.9) & 1 (0.2) & 0 (0.0) \\
OmniGen-v1 & 470 & 117 (24.9) & 260 (55.3) & 52 (11.1) & 41 (8.7) & 0 (0.0) \\
Qwen-Image-Edit-2511 & 470 & 115 (24.5) & 321 (68.3) & 29 (6.2) & 5 (1.1) & 0 (0.0) \\
BAGEL-7B-MoT & 470 & 117 (24.9) & 296 (63.0) & 18 (3.8) & 39 (8.3) & 0 (0.0) \\
\midrule
\rowcolor{black!7}
\textbf{All model--sample records} & 5,170 & 1779 (34.4) & 2985 (57.7) & 287 (5.6) & 117 (2.3) & 2 ($<0.1$) \\
\bottomrule
\end{tabular}
\end{table}

\begin{table}[H]
\centering
\scriptsize
\setlength{\tabcolsep}{3.8pt}
\renewcommand{\arraystretch}{1.20}
\caption{Failure attribution by SpatialGen-Bench task across the visual-answering systems. Each cell reports count (row percentage). Task totals differ because the benchmark contains 25--40 samples per task.}
\label{tab:failure-by-task}
\begin{tabular}{@{\hspace{5pt}}l r r r r r r@{\hspace{5pt}}}
\toprule
\textbf{Task} & \textbf{$N$} & \makecell{\textbf{Correct}\\\textbf{prediction}} & \makecell{\textbf{Incorrect}\\\textbf{prediction}} & \makecell{\textbf{Protocol}\\\textbf{noncompliance}} & \makecell{\textbf{Parser}\\\textbf{failure}} & \makecell{\textbf{Generation}\\\textbf{failure}} \\
\midrule
\rowcolor{black!5}\multicolumn{7}{l}{\textbf{Perception}} \\
Counting & 440 & 159 (36.1) & 281 (63.9) & 0 (0.0) & 0 (0.0) & 0 (0.0) \\
Depth & 330 & 225 (68.2) & 103 (31.2) & 2 (0.6) & 0 (0.0) & 0 (0.0) \\
Orientation & 440 & 139 (31.6) & 288 (65.5) & 13 (3.0) & 0 (0.0) & 0 (0.0) \\
Size & 385 & 191 (49.6) & 88 (22.9) & 105 (27.3) & 0 (0.0) & 1 (0.3) \\
\addlinespace[0.25em]
\rowcolor{black!5}\multicolumn{7}{l}{\textbf{Understanding}} \\
Relationship & 385 & 164 (42.6) & 221 (57.4) & 0 (0.0) & 0 (0.0) & 0 (0.0) \\
Perspective & 385 & 69 (17.9) & 309 (80.3) & 6 (1.6) & 0 (0.0) & 1 (0.3) \\
Mental modeling & 385 & 128 (33.2) & 249 (64.7) & 8 (2.1) & 0 (0.0) & 0 (0.0) \\
Grounding & 385 & 151 (39.2) & 193 (50.1) & 0 (0.0) & 41 (10.6) & 0 (0.0) \\
\addlinespace[0.25em]
\rowcolor{black!5}\multicolumn{7}{l}{\textbf{Reasoning}} \\
Multi-hop & 330 & 72 (21.8) & 253 (76.7) & 5 (1.5) & 0 (0.0) & 0 (0.0) \\
Prediction & 275 & 27 (9.8) & 181 (65.8) & 0 (0.0) & 67 (24.4) & 0 (0.0) \\
Feasibility & 385 & 117 (30.4) & 120 (31.2) & 148 (38.4) & 0 (0.0) & 0 (0.0) \\
\addlinespace[0.25em]
\rowcolor{black!5}\multicolumn{7}{l}{\textbf{Interaction}} \\
Affordance & 385 & 74 (19.2) & 311 (80.8) & 0 (0.0) & 0 (0.0) & 0 (0.0) \\
Navigation & 330 & 116 (35.2) & 214 (64.8) & 0 (0.0) & 0 (0.0) & 0 (0.0) \\
Trajectory & 330 & 147 (44.5) & 174 (52.7) & 0 (0.0) & 9 (2.7) & 0 (0.0) \\
\bottomrule
\end{tabular}
\end{table}

\subsection{Parsed-Output Sensitivity}
\label{app:parsed-output-sensitivity}

The fixed-denominator scores in the main results treat every invalid output as zero so that all systems are evaluated on the same items. Table~\ref{tab:parsed-output-scores} conditions each model--task score on outputs that yield a prediction consumable by the native metric. GPT Image 2 and Nano Banana 2 remain first and second. The largest upward shifts are SenseNova-Vision-7B-MoT (7$\rightarrow$3) and OmniGen-v1 (9$\rightarrow$6), while JoyAI-Image (5$\rightarrow$7) and FLUX.2 [klein] 9B (6$\rightarrow$8) move down. These changes reflect different retained subsets rather than improved capability, so the conditional table is diagnostic rather than a replacement leaderboard.

\begin{landscape}
\begin{center}
\captionsetup{hypcap=false}
\captionof{table}{Visual-answering results after filtering invalid parses. Scorable reports the percentage of outputs for which the assigned parser returns a prediction accepted by the native task metric. All-output is the official fixed-denominator score; Scorable-only excludes parse failures within each task before macro-aggregation.}
\label{tab:parsed-output-scores}
\scriptsize
\setlength{\tabcolsep}{2.6pt}
\renewcommand{\arraystretch}{1.30}

\begin{tabularx}{0.995\linewidth}{@{\hspace{4pt}}l*{11}{Y}@{\hspace{4pt}}}
\toprule
\multirow{2}{*}{\textbf{Model}} &
\multicolumn{4}{c}{\textbf{Summary}} &
\multicolumn{4}{c}{\textbf{Perception}} &
\multicolumn{3}{c}{\textbf{Understanding}} \\
\cmidrule(lr){2-5}\cmidrule(lr){6-9}\cmidrule(lr){10-12}
& \makecell{\textbf{Scorable}\\\textbf{(\%)}} & \makecell{\textbf{All-output}\\\textbf{score}} & \makecell{\textbf{Scorable-only}\\\textbf{score}} & \makecell{\textbf{Rank}\\\textbf{shift}}
& \textbf{Cnt.} & \textbf{Dep.} & \textbf{Ori.} & \textbf{Size}
& \textbf{Rel.} & \textbf{Pers.} & \textbf{Mental} \\
\midrule
GPT Image 2 & 98.7 & 54.49 & 55.02 & 1$\rightarrow$1 & 82.5 & 63.3 & 52.5 & 80.0 & 45.7 & 41.2 & 25.7 \\
\rowcolor{gray!6}Nano Banana 2 & 97.4 & 48.63 & 49.78 & 2$\rightarrow$2 & 82.5 & 86.7 & 64.1 & 65.6 & 45.7 & 25.7 & 31.4 \\
GPT-5 Image Mini & 97.0 & 40.00 & 41.09 & 3$\rightarrow$4 & 52.5 & 73.3 & 79.5 & 57.6 & 42.9 & 29.4 & 33.3 \\
\rowcolor{gray!6}Seedream 4.5 & 94.9 & 38.13 & 40.87 & 4$\rightarrow$5 & 30.0 & 90.0 & 40.0 & 59.3 & 45.7 & 20.0 & 28.1 \\
JoyAI-Image & 96.6 & 35.28 & 37.34 & 5$\rightarrow$7 & 55.0 & 53.3 & 5.0 & 76.0 & 40.0 & 17.1 & 25.7 \\
\rowcolor{gray!6}FLUX.2 [klein] 9B & 94.7 & 33.58 & 37.15 & 6$\rightarrow$8 & 25.0 & 82.8 & 20.0 & 69.7 & 40.0 & 8.6 & 45.7 \\
SenseNova-Vision-7B-MoT & 78.5 & 32.70 & 45.11 & 7$\rightarrow$3 & 22.5 & 83.3 & 14.3 & 80.0 & 48.6 & 9.4 & 37.1 \\
\rowcolor{gray!6}FLUX.2 [klein] 4B & 94.9 & 28.84 & 30.78 & 8$\rightarrow$9 & 20.0 & 48.3 & 32.5 & 75.0 & 40.0 & 14.3 & 31.4 \\
OmniGen-v1 & 80.2 & 26.72 & 38.32 & 9$\rightarrow$6 & 7.5 & 93.3 & 10.0 & 75.0 & 48.6 & 5.7 & 42.9 \\
\rowcolor{gray!6}Qwen-Image-Edit-2511 & 92.8 & 26.44 & 29.64 & 10$\rightarrow$10 & 10.0 & 43.3 & 17.6 & 73.1 & 31.4 & 15.2 & 34.3 \\
BAGEL-7B-MoT & 87.9 & 25.44 & 27.62 & 11$\rightarrow$11 & 10.0 & 36.7 & 20.0 & 55.6 & 40.0 & 14.3 & 37.1 \\
\bottomrule
\end{tabularx}

\vspace{0.75em}

\begin{tabularx}{0.98\linewidth}{@{\hspace{4pt}}l*{7}{Y}@{\hspace{4pt}}}
\toprule
\multirow{2}{*}{\textbf{Model}} &
\multicolumn{1}{c}{\textbf{Understanding}} &
\multicolumn{3}{c}{\textbf{Reasoning}} &
\multicolumn{3}{c}{\textbf{Interaction}} \\
\cmidrule(lr){2-2}\cmidrule(lr){3-5}\cmidrule(lr){6-8}
& \textbf{Ground} & \textbf{M-Hop} & \textbf{Pred.} & \textbf{Feas.}
& \textbf{Aff.} & \textbf{Nav.} & \textbf{Traj.} \\
\midrule
GPT Image 2 & 79.4 & 40.0 & 54.1 & 35.5 & 60.4 & 40.0 & 70.0 \\
\rowcolor{gray!6}Nano Banana 2 & 65.7 & 17.9 & 21.0 & 44.8 & 42.5 & 40.0 & 63.3 \\
GPT-5 Image Mini & 34.3 & 10.7 & 30.7 & 43.8 & 14.1 & 36.7 & 36.7 \\
\rowcolor{gray!6}Seedream 4.5 & 52.9 & 10.0 & 36.1 & 60.9 & 9.1 & 36.7 & 53.3 \\
JoyAI-Image & 57.1 & 10.0 & 14.2 & 41.4 & 34.5 & 33.3 & 60.0 \\
\rowcolor{gray!6}FLUX.2 [klein] 9B & 11.8 & 23.3 & 16.5 & 71.4 & 28.6 & 36.7 & 40.0 \\
SenseNova-Vision-7B-MoT & 91.2 & 33.3 & 100.0 & 0.0 & 51.1 & 20.0 & 40.7 \\
\rowcolor{gray!6}FLUX.2 [klein] 4B & 23.5 & 6.9 & 18.5 & 35.3 & 11.8 & 36.7 & 36.7 \\
OmniGen-v1 & 5.9 & 33.3 & 61.3 & 71.4 & 5.6 & 40.0 & 36.0 \\
\rowcolor{gray!6}Qwen-Image-Edit-2511 & 19.4 & 23.3 & 28.6 & 56.5 & 4.7 & 33.3 & 24.1 \\
BAGEL-7B-MoT & 4.8 & 33.3 & 0.0 & 56.0 & 5.6 & 33.3 & 40.0 \\
\bottomrule
\end{tabularx}

\vspace{0.45em}
\begin{minipage}{0.98\linewidth}
\footnotesize
A scorable prediction may be correct or incorrect. Outputs that do not yield a metric-consumable prediction are omitted only from this diagnostic; because its denominators differ by model and task, the Scorable-only score is not an alternative leaderboard score. A task with no scorable prediction remains zero rather than disappearing from the macro-average.
\end{minipage}
\end{center}
\end{landscape}

\section{Qualitative Interface Complementarity}
\label{app:qualitative-complementarity}

\begin{figure}[H]
    \centering
    \begin{minipage}[c]{0.21\textwidth}
        \centering
        \includegraphics[width=\linewidth]{img/interface_rescues/depth_consensus_input.jpg}\\[-1pt]
        {\scriptsize Source input}
    \end{minipage}
    \hspace{0.07\textwidth}
    \begin{minipage}[c]{0.21\textwidth}
        \centering
        \includegraphics[width=\linewidth]{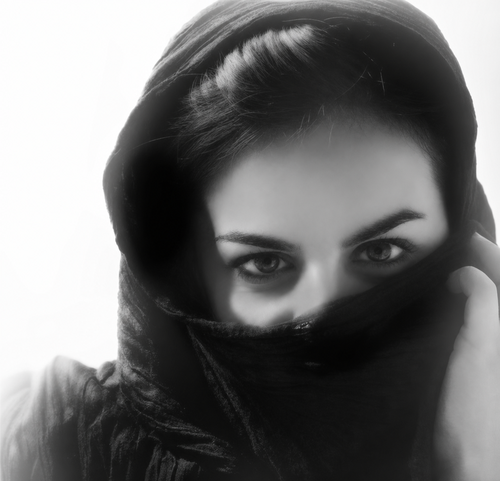}\\[-1pt]
        {\scriptsize GPT Image 2 response}
    \end{minipage}
    \caption[\textbf{Depth qualitative example.}]{\textbf{Depth.} \textbf{Question:} Which point is closer to the camera? \textbf{Options:} (A) A is closer; (B) B is closer. \textbf{GPT-5.4:} A. \textbf{Ground truth:} B. The generated depth field supports B at the benchmark coordinates; GPT Image 2 and all 11 visual systems pass this item.}
    \label{fig:qualitative-depth}
\end{figure}

\begin{figure}[H]
    \centering
    \begin{minipage}[c]{0.21\textwidth}
        \centering
        \includegraphics[width=\linewidth]{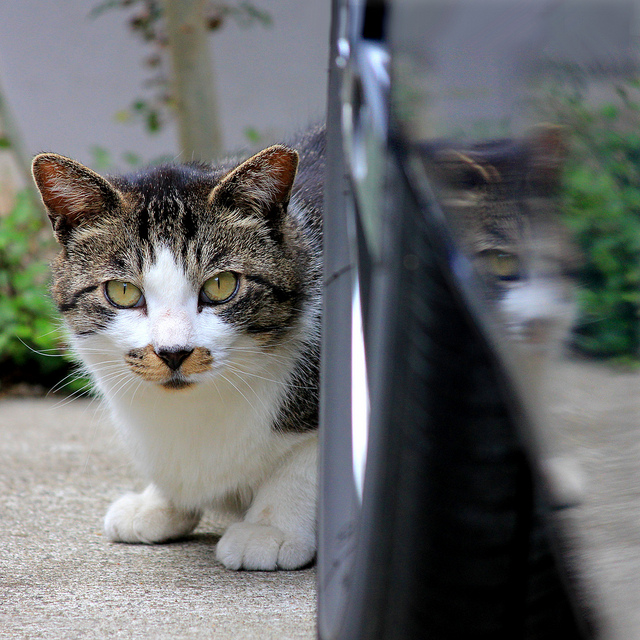}\\[-1pt]
        {\scriptsize Source input}
    \end{minipage}
    \hspace{0.07\textwidth}
    \begin{minipage}[c]{0.21\textwidth}
        \centering
        \includegraphics[width=\linewidth]{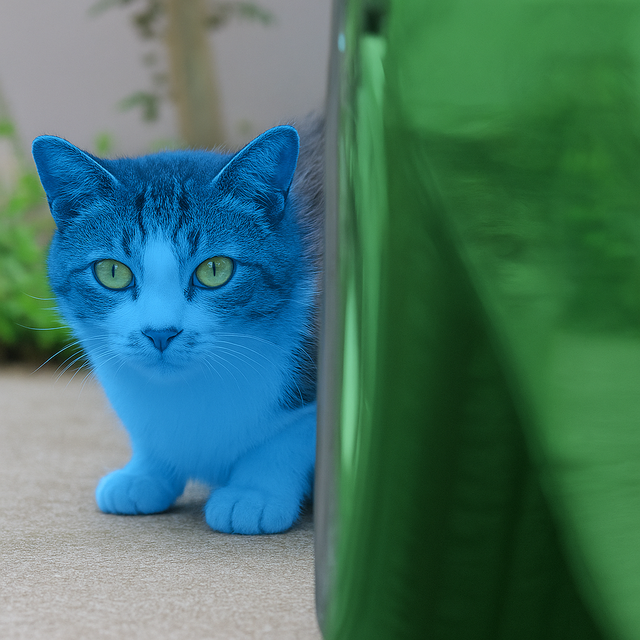}\\[-1pt]
        {\scriptsize GPT-5 Image Mini response}
    \end{minipage}
    \caption[\textbf{Relationship qualitative example.}]{\textbf{Relationship.} \textbf{Question:} Is the following statement true or false: The car is in front of the cat. \textbf{Options:} True; False. \textbf{GPT-5.4:} False. \textbf{Ground truth:} True. GPT-5 Image Mini marks the cat and foreground vehicle as distinct, parser-readable regions; 8 of 11 visual systems pass.}
    \label{fig:qualitative-relationship}
\end{figure}

\begin{figure}[H]
    \centering
    \begin{minipage}[c]{0.21\textwidth}
        \centering
        \includegraphics[width=\linewidth]{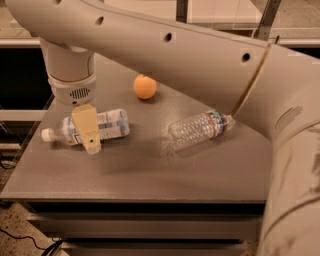}\\[-1pt]
        {\scriptsize Source input}
    \end{minipage}
    \hspace{0.07\textwidth}
    \begin{minipage}[c]{0.21\textwidth}
        \centering
        \includegraphics[width=\linewidth]{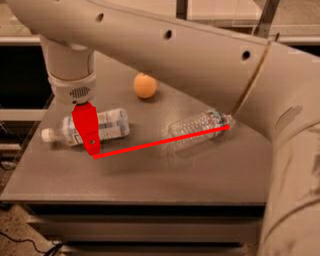}\\[-1pt]
        {\scriptsize JoyAI-Image response}
    \end{minipage}
    \caption[\textbf{Trajectory qualitative example.}]{\textbf{Trajectory.} \textbf{Question:} You are a Franka robot using the end effector control. The task is pick blue plastic bottle. \textbf{Answer space:} a five-point trajectory polyline in normalized image coordinates (no discrete options). \textbf{GPT-5.4:} \((0.266,0.465)\rightarrow(0.330,0.495)\rightarrow(0.395,0.525)\rightarrow(0.455,0.550)\rightarrow(0.510,0.575)\). \textbf{Ground truth:} \((0.266,0.465)\rightarrow(0.278,0.465)\rightarrow(0.250,0.422)\rightarrow(0.209,0.348)\rightarrow(0.200,0.320)\). The JoyAI-Image path passes the DFD criterion; 7 of 11 visual systems pass.}
    \label{fig:qualitative-trajectory}
\end{figure}

Figures~\ref{fig:qualitative-depth}, \ref{fig:qualitative-relationship}, and~\ref{fig:qualitative-trajectory} complement the paired outcome counts in Table~\ref{tab:interface-complementarity}. Each pairs an incorrect response from the named text-output baseline with a displayed visual response that passes the unchanged task metric. The captions record the complete question, answer space, text response, and benchmark answer. Source inputs and generated responses are shown without cropping or aspect-ratio normalization.

\begin{figure}[H]
    \centering
    \begin{minipage}[c]{0.26\textwidth}
        \centering
        \includegraphics[width=\linewidth]{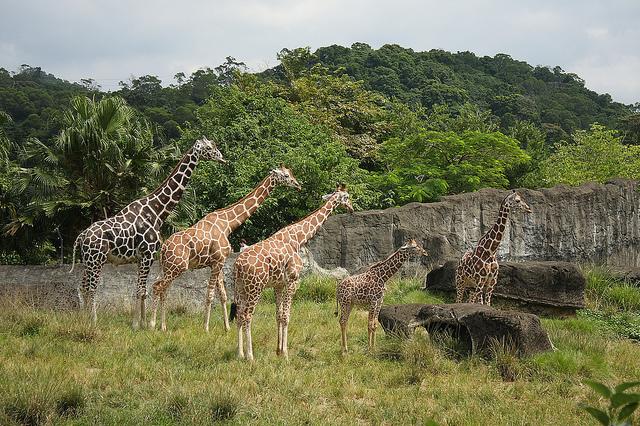}\\[-1pt]
        {\scriptsize Source input}
    \end{minipage}
    \hspace{0.05\textwidth}
    \begin{minipage}[c]{0.26\textwidth}
        \centering
        \includegraphics[width=\linewidth]{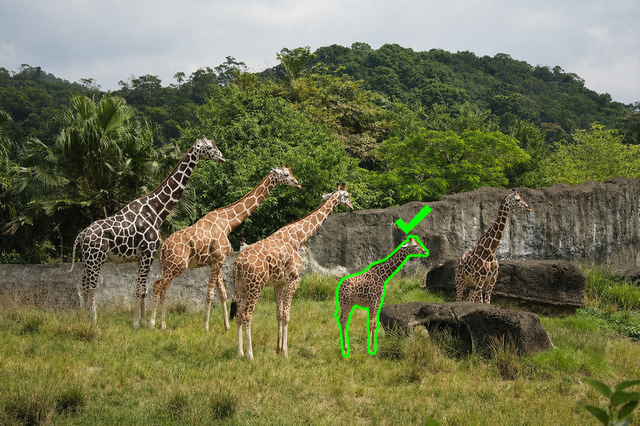}\\[-1pt]
        {\scriptsize GPT Image 2 response}
    \end{minipage}
    \caption[\textbf{Size qualitative example.}]{\textbf{Size.} \textbf{Question:} Is the second giraffe from the right the shortest? \textbf{Options:} (A) no; (B) yes. \textbf{GPT-5.4:} A (no). \textbf{Ground truth:} B (yes). GPT Image 2 outlines the queried giraffe and gives an affirmative visual mark; 9 of 11 visual systems pass.}
    \label{fig:qualitative-size}
\end{figure}

\begin{figure}[H]
    \centering
    \begin{minipage}[c]{0.30\linewidth}
        \centering
        \includegraphics[width=\linewidth]{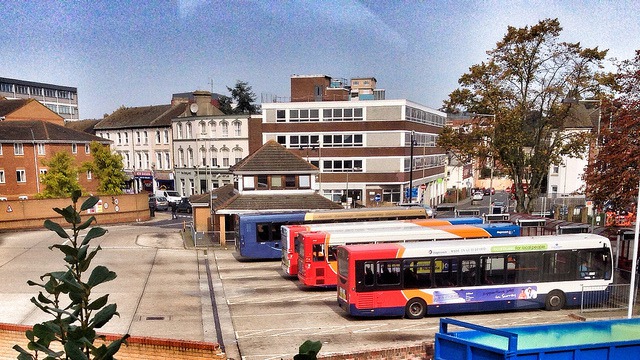}\\[-1pt]
        {\scriptsize Source input}
    \end{minipage}
    \hfill
    \begin{minipage}[c]{0.30\linewidth}
        \centering
        \includegraphics[width=\linewidth]{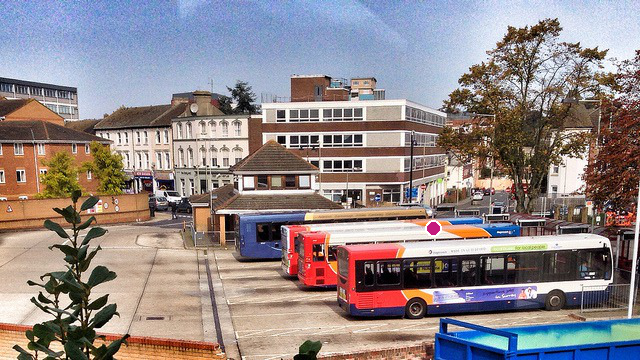}\\[-1pt]
        {\scriptsize GPT-5.4 text point (overlaid)}
    \end{minipage}
    \hfill
    \begin{minipage}[c]{0.30\linewidth}
        \centering
        \includegraphics[width=\linewidth]{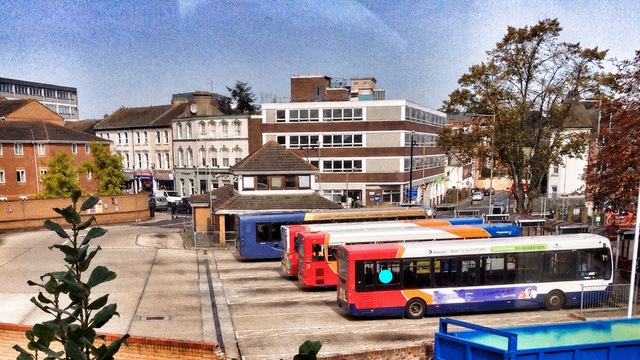}\\[-1pt]
        {\scriptsize GPT Image 2 response}
    \end{minipage}
    \caption[\textbf{Spatial grounding qualitative example.}]{\textbf{Spatial Grounding.} \textbf{Question:} Locate the object described as \textit{a bus in front of others}. Return one normalized point inside the target object. \textbf{Answer space:} a normalized point (no discrete options). The ground-truth target is the RefCOCOg segmentation mask of the foreground bus. GPT-5.4 predicts \((0.677,0.634)\), shown in magenta on a copy of the source image; the point falls just outside the target mask. The parser reads \((0.603,0.768)\) from the GPT Image 2 response, placing the cyan marker inside the mask; 10 of 11 visual systems pass.}
    \label{fig:qualitative-grounding}
\end{figure}

\begin{figure}[H]
    \centering
    \begin{minipage}[c]{0.26\textwidth}
        \centering
        \includegraphics[width=\linewidth]{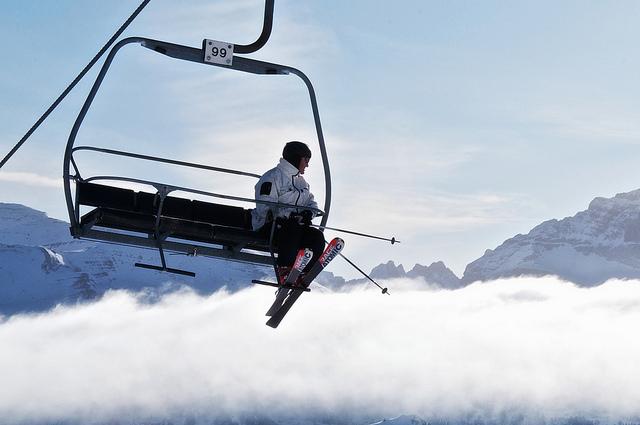}\\[-1pt]
        {\scriptsize Source input}
    \end{minipage}
    \hspace{0.05\textwidth}
    \begin{minipage}[c]{0.26\textwidth}
        \centering
        \includegraphics[width=\linewidth]{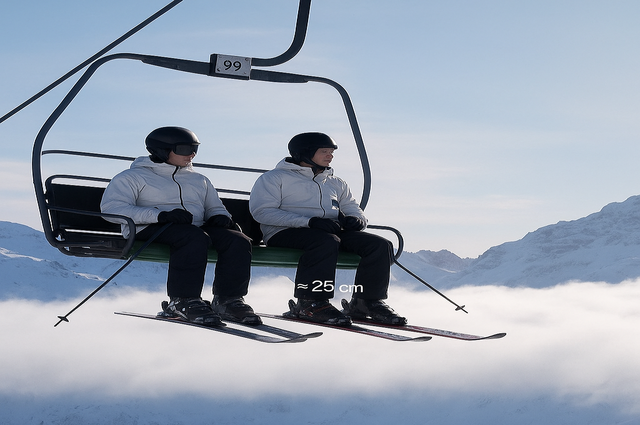}\\[-1pt]
        {\scriptsize GPT-5 Image Mini response}
    \end{minipage}
    \caption[\textbf{Geometric feasibility qualitative example.}]{\textbf{Geometric Feasibility.} \textbf{Question:} Is there enough space for another person to sit in the ski lift? \textbf{Options:} (A) no; (B) yes. \textbf{RoboBrain2.5-8B-NV:} A (no). \textbf{Ground truth:} B (yes). GPT-5 Image Mini places a second passenger on the same bench with visible remaining clearance; 8 of 11 visual systems pass.}
    \label{fig:qualitative-feasibility}
\end{figure}

\begin{figure}[H]
    \centering
    \begin{minipage}[c]{0.26\textwidth}
        \centering
        \includegraphics[width=\linewidth]{img/interface_rescues/orientation_consensus_input.jpg}\\[-1pt]
        {\scriptsize Source input}
    \end{minipage}
    \hspace{0.05\textwidth}
    \begin{minipage}[c]{0.26\textwidth}
        \centering
        \includegraphics[width=\linewidth]{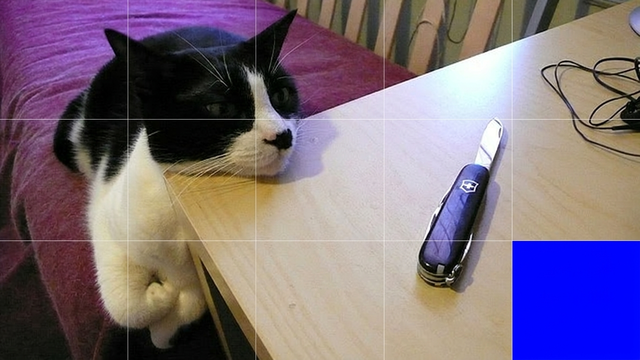}\\[-1pt]
        {\scriptsize Nano Banana 2 response}
    \end{minipage}
    \caption[\textbf{Orientation qualitative example.}]{\textbf{Orientation.} \textbf{Question:} From the camera's perspective, which direction is the cat facing? \textbf{Options:} (A) front; (B) front-right; (C) right; (D) back-right; (E) back; (F) back-left; (G) left; (H) front-left. \textbf{GPT-5.4:} C (right). \textbf{Ground truth:} B (front-right). Nano Banana 2 fills the bottom-right grid cell, which the deterministic parser maps to front-right; 7 of 11 visual systems pass.}
    \label{fig:qualitative-orientation}
\end{figure}

\noindent\textbf{Additional task examples.}
Figures~\ref{fig:qualitative-size}, \ref{fig:qualitative-grounding}, \ref{fig:qualitative-feasibility}, and~\ref{fig:qualitative-orientation} cover the released Size, Spatial Grounding, Geometric Feasibility, and Orientation tasks; the final case illustrates the direction-grid protocol.

\section{Experimental Settings and Scoring}
\label{app:evaluation_settings}

This section specifies the evaluation interface used after the benchmark split and visual protocols are fixed. It covers text-output prompts, model runtime settings, and score aggregation. Protocol constraints and the released generation-prompt locations are documented in Appendix~\ref{app:prompt_parser_protocols}.

\subsection{Human Evaluation}
\label{app:human_reference}

The Human reference covers the full benchmark and is scored with the same task-specific metrics used for model outputs. Capability and Overall values use the same macro-aggregation in Eq.~\ref{eq:app_score_aggregation}. This evaluation supplies the Human row in the result tables and is separate from the source-data quality audit in Appendix~\ref{app:source_data_quality_assessment}.

\subsection{Text-Output VLM Prompt Formats}
\label{app:vlm_prompts}

Text-output VLMs are prompted to answer in the original benchmark answer space without chain-of-thought or explanatory prose. The response constraint is matched to the target answer type: a single integer for counting, A/B for relative depth and object size, a normalized direction label for orientation and perspective, \texttt{true}/\texttt{false} for relation and feasibility decisions, an option label for multiple-choice tasks, normalized points for spatial grounding, and the structured textual form required by the corresponding original evaluator for mask or trajectory targets. This keeps the text baseline and the visual protocol scored against the same target format.

\subsection{Model and Scoring Settings}
\label{app:model_scoring_settings}

Local experiments use one NVIDIA RTX PRO 6000 GPU; API systems use the listed provider-compatible endpoints. Local VLM decoding is deterministic with batch size one. Each visual system produces one protocol-constrained image per sample, with no correctness retries or manual selection; cached outputs are reused for rescoring. Editing requests preserve source aspect ratio, target approximately 1K resolution when supported, and are resized to the source resolution before parsing. Missing or invalid outputs receive the declared invalid-output score. Table~\ref{tab:appendix_model_settings} records the evaluated model families and runtime settings.

\begingroup
\footnotesize
\setlength{\tabcolsep}{4pt}
\renewcommand{\arraystretch}{1.12}
\setlength{\LTleft}{\fill}
\setlength{\LTright}{\fill}
\begin{longtable}{@{}L{0.32\textwidth} L{0.64\textwidth}@{}}
\caption{Complete model list and runtime settings for SpatialGen-Bench evaluation.}
\label{tab:appendix_model_settings}\\
\toprule
\textbf{Model} & \textbf{Runtime setting} \\
\midrule
\endfirsthead
\multicolumn{2}{@{}l}{\footnotesize\textit{Table~\thetable\ continued.}} \\
\toprule
\textbf{Model} & \textbf{Runtime setting} \\
\midrule
\endhead
\bottomrule
\endfoot
\bottomrule
\endlastfoot

\rowcolor{red!4}
\multicolumn{2}{@{}l}{
\textbf{Text-output VLMs} \quad
\textit{Direct answers in the original benchmark answer space.}
} \\
Qwen3-VL-8B~\citep{bai2025qwen3vltechnicalreport}
& Local inference; deterministic decoding; batch size 1; max 512 new tokens. \\
InternVL3.5-8B~\citep{wang2025internvl35advancingopensourcemultimodal}
& Local BF16 inference; dynamic image tiling; deterministic decoding. \\
LLaVA-OV-1.5-8B~\citep{an2025llavaonevision15fullyopenframework}
& Local BF16 inference; deterministic decoding. \\
GLM-4.6V-Flash~\citep{vteam2025glm45vglm41vthinkingversatilemultimodal}
& Local BF16 inference; deterministic decoding. \\
GPT-5.4
& OpenAI-compatible chat API; request model \path{gpt-5.4}; direct structured answer; deterministic request settings where supported. \\
Cosmos3-Nano; RynnBrain-8B; RoboBrain2.5-8B-NV~\citep{nvidia2026cosmos3,dang2026rynnbrain,tan2026robobrain25}
& Local multimodal inference; model-native preprocessing; deterministic decoding; batch size 1. \\
SpaceR; Cambrian-S-7B; GEM-2B; SpatialThinker-3B~\citep{ouyang2025spacer,yang2025cambrian,zhao2026gem,batra2025spatialthinker}; SpaceOm
& Local spatial-specialized VLM inference; released checkpoint and processor; deterministic decoding. \\
VLM-3R-7B; SpatialRGPT-8B; Spatial-MLLM; SpatialBot-3B~\citep{fan2025vlm3r,cheng2024spatialrgpt,wu2025spatial,cai2025spatialbot}
& Local spatial-specialized VLM inference; released checkpoint and task-compatible prompt format. \\
SenseNova-Vision-7B-MoT; Janus-Pro-7B; Janus-1.3B~\citep{han2026sensenovavision,chen2025janus,wu2024janus}
& Local unified-model text interface; deterministic decoding; batch size 1. \\

\rowcolor{gray!4}
\multicolumn{2}{@{}p{\dimexpr0.98\textwidth-2\tabcolsep\relax}@{}}{
\textit{Output processing:} raw responses are normalized to the original answer format; invalid formats are scored as incorrect.
} \\

\addlinespace[0.35em]
\midrule

\rowcolor{green!5}
\multicolumn{2}{@{}l}{
\textbf{Image-output generative models} \quad
\textit{Protocol-constrained visual responses parsed back to structured predictions.}
} \\
JoyAI-Image~\citep{joyfuture2026joyaiimage}
& Local image editing; seed 42; 50 inference steps; guidance 5.0; base size 1024. \\
FLUX.2 [klein] 4B
& Local image editing; seed 42; 4 inference steps; guidance 1.0; max side 1024. \\
FLUX.2 [klein] 9B
& Local image editing with the 9B checkpoint; one protocol-constrained image per sample. \\
SenseNova-Vision-7B-MoT; BAGEL-7B-MoT~\citep{han2026sensenovavision,deng2025emerging}
& Local unified-model visual interface; one protocol-constrained image per sample. \\
Qwen-Image-Edit-2511~\citep{wu2025qwenimage}
& Local BF16 image editing; model offload; seed 0; 40 steps; true CFG 4.0; max side 1024. \\
OmniGen-v1~\citep{xiao2024omnigen}
& Local BF16 image editing; seed 42; 50 steps; guidance 2.5; image guidance 1.6; max side 1024. \\
Seedream 4.5
& Provider-compatible image API; request route \path{bytedance-seed/seedream-4.5}; one protocol-constrained image per sample. \\
GPT-5 Image Mini
& OpenRouter image route \path{openai/gpt-5-image-mini}; provider-composed request alias; one protocol-constrained image per sample. \\
Nano Banana 2
& Provider-compatible image route \path{google/gemini-3.1-flash-image-preview} (dated alias \path{google/gemini-3.1-flash-image-preview-20260226}); one image per sample. \\
GPT Image 2
& OpenAI-compatible image-edit API; request model \path{gpt-image-2}; returned alias \path{gpt-image-2-codex}; source image plus protocol prompt; one image per sample. \\

\rowcolor{gray!4}
\multicolumn{2}{@{}p{\dimexpr0.98\textwidth-2\tabcolsep\relax}@{}}{
\textit{Output processing:} one output image per sample; outputs are resized to the source resolution if needed and parsed automatically with the assigned protocol. Missing, unreadable, or unparseable outputs are marked invalid.
} \\

\end{longtable}
\endgroup

\subsection{Score Aggregation}
\label{app:score_aggregation}

All parsed predictions are first converted into per-sample scores $s_i\in[0,1]$. Counting, relative-depth choice, orientation, object size, relationship, perspective, mental modeling, multihop reasoning, geometric feasibility, and navigation use exact-match accuracy after answer normalization. The remaining tasks use the task-specific scores below; all reported numbers are percentages.

For spatial grounding, the parser recovers one normalized point $\hat p$ from the cyan marker and the source annotation provides a binary target mask $M$. The per-sample score preserves the source point-in-mask contract:
\begin{equation}
s_{\mathrm{ground}}(\hat p,M)=\mathbf{1}[M(\hat p)>0].
\label{eq:app_grounding_score}
\end{equation}

For affordance grounding, $M$ denotes the annotated target region. The prediction $\hat M$ is the parsed binary mask for image-output models, or the set of predicted points rasterized to image coordinates for text-output VLMs. The per-sample score is precision:
\begin{equation}
s_{\mathrm{aff}}(\hat M,M)
=
\frac{|\hat M \cap M|}{|\hat M|+\epsilon},
\label{eq:app_affordance_score}
\end{equation}
where $\epsilon$ prevents division by zero for empty predictions. This score measures how much of the predicted actionable support lies inside the annotated region; the reported benchmark score uses the mean precision across samples.

For ball-state prediction, $\hat y$ and $y$ are the predicted and ground-truth hole indices, and $H$ is the number of candidate holes. Equation~\ref{eq:app_prediction_score} assigns partial credit only to adjacent-hole predictions:
\begin{equation}
s_{\mathrm{pred}}(\hat y,y,H)=
\begin{cases}
1, & |\hat y-y|=0,\\
0.8-\frac{1}{H}, & |\hat y-y|=1,\\
0, & |\hat y-y|\ge 2,
\end{cases}
\label{eq:app_prediction_score}
\end{equation}
Exact predictions receive full credit, adjacent-hole errors receive a smaller score adjusted by $H$, and larger errors receive zero.

For trajectory planning, let $\hat P=(\hat p_1,\ldots,\hat p_n)$ and $P=(p_1,\ldots,p_m)$ be the parsed and ground-truth point sequences after normalization to image coordinates. We compute Discrete Fr\'echet Distance and threshold it as in Eq.~\ref{eq:app_trajectory_score}:
\begin{equation}
d_{\mathrm{DFD}}(\hat P,P)
=
\min_{C}
\max_{(\hat p,p)\in C}
\|\hat p-p\|_2,
\qquad
s_{\mathrm{traj}}=\mathbf{1}[d_{\mathrm{DFD}}(\hat P,P)<0.4],
\label{eq:app_trajectory_score}
\end{equation}
where $C$ ranges over monotone couplings between the two point sequences. The trajectory score is therefore a binary success indicator for geometric path agreement.

Equation~\ref{eq:app_score_aggregation} defines the aggregation rule. For task $t$, $N_t$ is the number of samples; for capability $c$, $\mathcal{T}_c$ is its subtask set:
\begin{equation}
S_t=\frac{1}{N_t}\sum_{i=1}^{N_t}s_i,\qquad
S_c=\frac{1}{|\mathcal{T}_c|}\sum_{t\in\mathcal{T}_c}S_t,\qquad
S_{\mathrm{overall}}=\frac{1}{14}\sum_{t=1}^{14}S_t.
\label{eq:app_score_aggregation}
\end{equation}
Task scores are averaged over samples, capability scores are macro-averaged over subtasks within each capability, and the overall score is macro-averaged over all 14 subtasks. This prevents larger source tasks from dominating the final benchmark score.

\subsection{Statistical Uncertainty}
\label{app:statistical_uncertainty}

We estimate sample-selection uncertainty on the complete benchmark split. Each of 10,000 replicates resamples examples with replacement within every task, recomputes task means, and then applies the same task-macro aggregation as the leaderboard. Intervals are percentile 95\% confidence intervals for fixed cached outputs; they do not capture generation stochasticity or provider-side model updates.

\clearpage
\begin{landscape}
\begin{center}
\captionsetup{hypcap=false}
\captionof{table}{Bootstrap 95\% confidence intervals for all model-interface systems. Entries are point estimates [2.5th, 97.5th percentile].}
\label{tab:appendix_bootstrap_full}
\footnotesize
\setlength{\tabcolsep}{2.8pt}
\renewcommand{\arraystretch}{1.06}
\begin{tabular*}{0.98\linewidth}{@{\extracolsep{\fill}}l*{5}{c}@{}}
\toprule
\textbf{Model} & \textbf{Overall} & \textbf{Perception} & \textbf{Understanding} & \textbf{Reasoning} & \textbf{Interaction} \\
\midrule
\rowcolor{blue!5}\multicolumn{6}{@{}l}{\textbf{Text Answering}} \\
GPT-5.4 & 61.04 [57.12, 64.94] & 74.61 [67.26, 81.58] & 44.29 [35.71, 52.86] & 56.64 [47.42, 65.42] & 69.70 [62.13, 76.71] \\
Cosmos3-Nano & 60.17 [56.45, 63.86] & 76.82 [70.15, 82.98] & 45.00 [37.86, 52.14] & 55.20 [46.22, 64.24] & 63.15 [56.09, 70.31] \\
RynnBrain-8B & 51.15 [47.27, 54.93] & 67.05 [60.00, 73.81] & 45.71 [38.57, 52.86] & 44.21 [34.62, 53.60] & 44.13 [36.19, 52.54] \\
RoboBrain2.5-8B-NV & 47.43 [43.87, 51.10] & 58.39 [51.93, 64.61] & 34.29 [27.86, 40.71] & 47.76 [38.61, 56.79] & 50.03 [42.01, 58.10] \\
SenseNova-Vision-7B-MoT & 46.78 [42.63, 51.01] & 54.40 [46.93, 62.05] & 43.57 [36.43, 51.43] & 40.85 [31.42, 50.21] & 46.83 [37.62, 56.03] \\
InternVL3.5-8B & 46.25 [42.55, 50.05] & 51.70 [45.45, 57.92] & 36.43 [29.29, 43.57] & 62.45 [53.76, 71.14] & 35.87 [27.30, 44.76] \\
SpaceR & 45.84 [41.81, 49.84] & 61.04 [53.51, 68.51] & 31.43 [24.29, 38.57] & 53.24 [43.71, 62.50] & 37.38 [29.13, 45.63] \\
SpaceOm & 44.76 [40.68, 48.83] & 59.23 [51.64, 66.82] & 40.00 [32.86, 47.14] & 43.78 [34.79, 53.18] & 32.78 [24.21, 41.11] \\
LLaVA-OV-1.5-8B & 44.64 [40.68, 48.67] & 54.23 [47.50, 60.86] & 37.86 [30.00, 45.71] & 39.97 [30.91, 49.42] & 45.56 [37.30, 53.65] \\
SpatialThinker-3B & 44.62 [40.63, 48.73] & 57.65 [50.33, 64.91] & 41.43 [34.29, 49.29] & 39.75 [30.44, 49.16] & 36.37 [27.98, 44.92] \\
Qwen3-VL-8B & 44.50 [40.51, 48.47] & 59.79 [52.17, 67.35] & 37.14 [30.00, 44.29] & 47.00 [37.94, 55.95] & 31.43 [23.33, 39.68] \\
GEM-2B & 43.65 [39.57, 47.80] & 56.70 [48.84, 64.35] & 43.57 [35.71, 51.43] & 39.70 [30.96, 48.74] & 30.32 [22.22, 38.89] \\
Cambrian-S-7B & 42.55 [38.67, 46.42] & 68.01 [60.65, 75.18] & 21.43 [15.00, 28.57] & 47.09 [38.41, 55.97] & 32.22 [23.81, 40.95] \\
GLM-4.6V-Flash & 40.96 [37.33, 44.73] & 56.28 [49.70, 62.92] & 45.71 [37.86, 53.57] & 29.75 [22.70, 36.95] & 25.42 [17.98, 33.42] \\
VLM-3R-7B & 36.90 [33.21, 40.60] & 56.19 [49.46, 62.86] & 21.43 [15.00, 28.57] & 37.52 [29.10, 46.25] & 31.21 [22.69, 39.75] \\
Spatial-MLLM & 32.82 [29.20, 36.54] & 32.77 [27.53, 38.27] & 30.00 [22.86, 37.86] & 32.97 [24.55, 41.86] & 36.51 [27.94, 44.92] \\
SpatialRGPT-8B & 31.72 [28.05, 35.48] & 38.90 [32.38, 45.39] & 27.86 [20.71, 35.00] & 29.47 [21.43, 38.00] & 29.52 [22.06, 37.14] \\
Janus-Pro-7B & 31.58 [28.03, 35.21] & 39.20 [32.98, 45.09] & 26.43 [19.29, 33.57] & 35.75 [27.36, 44.09] & 24.13 [16.67, 31.75] \\
Janus-1.3B & 28.59 [25.08, 32.14] & 39.43 [32.26, 46.61] & 28.57 [22.14, 35.00] & 31.62 [23.66, 39.66] & 11.11 [5.56, 17.78] \\
SpatialBot-3B & 27.99 [24.41, 31.61] & 33.12 [26.85, 39.55] & 32.86 [25.00, 40.71] & 31.85 [23.28, 40.53] & 10.79 [5.40, 16.98] \\
\addlinespace[0.16em]
\rowcolor{green!5}\multicolumn{6}{@{}l}{\textbf{Visual Answering}} \\
GPT Image 2 & 54.49 [50.38, 58.62] & 69.58 [62.11, 76.76] & 47.14 [39.29, 55.00] & 41.83 [32.62, 51.25] & 56.80 [48.08, 65.80] \\
Nano Banana 2 & 48.63 [44.56, 52.59] & 72.92 [65.74, 79.64] & 42.14 [34.29, 50.00] & 24.92 [16.95, 33.40] & 48.61 [39.52, 58.01] \\
GPT-5 Image Mini & 40.00 [36.02, 43.95] & 64.40 [56.73, 71.90] & 33.57 [25.71, 41.43] & 26.89 [18.85, 35.09] & 29.13 [21.04, 37.52] \\
Seedream 4.5 & 38.13 [34.27, 42.00] & 51.43 [44.29, 58.69] & 35.71 [27.86, 43.57] & 28.69 [20.86, 36.78] & 33.04 [24.79, 41.41] \\
JoyAI-Image & 35.28 [31.40, 39.19] & 41.90 [34.37, 49.35] & 35.00 [27.86, 42.86] & 19.50 [12.07, 27.41] & 42.61 [33.71, 51.58] \\
FLUX.2 [klein] 9B & 33.58 [29.74, 37.39] & 47.68 [40.54, 54.64] & 26.43 [20.00, 33.57] & 22.80 [14.70, 31.00] & 35.09 [26.37, 44.31] \\
SenseNova-Vision-7B-MoT & 32.70 [29.29, 36.13] & 32.44 [26.52, 38.33] & 45.71 [39.29, 52.86] & 12.44 [6.67, 18.89] & 35.93 [27.54, 44.59] \\
FLUX.2 [klein] 4B & 28.84 [25.15, 32.69] & 41.93 [34.61, 49.52] & 27.14 [20.71, 34.29] & 14.11 [7.32, 21.33] & 28.37 [20.29, 36.91] \\
OmniGen-v1 & 26.72 [23.42, 30.18] & 29.85 [25.56, 34.43] & 25.00 [18.57, 31.43] & 26.36 [17.86, 35.37] & 25.21 [17.52, 33.22] \\
Qwen-Image-Edit-2511 & 26.44 [22.70, 30.22] & 30.65 [23.84, 37.77] & 24.29 [17.86, 31.43] & 29.69 [21.03, 38.66] & 20.47 [13.38, 28.25] \\
BAGEL-7B-MoT & 25.44 [21.85, 29.20] & 27.38 [20.57, 34.55] & 23.57 [17.14, 30.00] & 24.44 [16.98, 32.54] & 26.32 [18.02, 34.59] \\
\bottomrule
\end{tabular*}
\end{center}

\clearpage

\subsection{Complete Task-Level Results}
\label{app:complete_task_results}
\begin{center}
\captionsetup{hypcap=false}
\captionof{table}{
Complete task-level SpatialGen-Bench results (\%). Tasks are grouped by capability; all capability and overall aggregates appear in Table~\ref{tab:main-results-vertical}.
}
\label{tab:full-task-results}
\scriptsize
\setlength{\tabcolsep}{7.4pt}
\renewcommand{\arraystretch}{1.18}
\begin{tabular}{@{\hspace{4pt}}l*{14}{c}@{\hspace{4pt}}}
\toprule
\multirow{2}{*}{\textbf{Model}} &
\multicolumn{4}{c}{\textbf{Perception}} &
\multicolumn{4}{c}{\textbf{Understanding}} &
\multicolumn{3}{c}{\textbf{Reasoning}} &
\multicolumn{3}{c}{\textbf{Interaction}} \\
\cmidrule(lr){2-5}\cmidrule(lr){6-9}\cmidrule(lr){10-12}\cmidrule(lr){13-15}
& \textbf{Cnt.} & \textbf{Dep.} & \textbf{Ori.} & \textbf{Size}
& \textbf{Rel.} & \textbf{Pers.} & \textbf{Mental} & \textbf{Ground.}
& \textbf{M-Hop} & \textbf{Pred.} & \textbf{Feas.}
& \textbf{Aff.} & \textbf{Nav.} & \textbf{Traj.} \\
\midrule
\rowcolor{black!10}
\textbf{Human} & 100.00 & 100.00 & 92.50 & 97.14 & 88.57 & 91.43 & 74.29 & 88.57 & 63.33 & 79.00 & 91.43 & 89.48 & 83.33 & 90.00 \\
\midrule
\rowcolor{blue!5}\multicolumn{15}{l}{\textbf{\textit{Text Answering}}} \\
GPT-5.4 & 80.00 & 76.67 & 67.50 & 74.29 & 42.86 & 31.43 & 45.71 & 57.14 & 50.00 & 48.50 & 71.43 & 42.43 & 83.33 & 83.33 \\
Cosmos3-Nano & 55.00 & 83.33 & 77.50 & 91.43 & 22.86 & 37.14 & 34.29 & 85.71 & 46.67 & 61.80 & 57.14 & 76.13 & 23.33 & 90.00 \\
RynnBrain-8B & 70.00 & 80.00 & 32.50 & 85.71 & 37.14 & 22.86 & 40.00 & 82.86 & 43.33 & 35.00 & 54.29 & 85.71 & 26.67 & 20.00 \\
RoboBrain2.5-8B-NV & 55.00 & 80.00 & 10.00 & 88.57 & 25.71 & 20.00 & 11.43 & 80.00 & 33.33 & 47.10 & 62.86 & 80.08 & 13.33 & 56.67 \\
SenseNova-Vision-7B-MoT & 67.50 & 33.33 & 42.50 & 74.29 & 42.86 & 34.29 & 22.86 & 74.29 & 30.00 & 49.70 & 42.86 & 63.81 & 26.67 & 50.00 \\
InternVL3.5-8B & 32.50 & 70.00 & 10.00 & 94.29 & 20.00 & 20.00 & 28.57 & 77.14 & 56.67 & 76.40 & 54.29 & 40.95 & 26.67 & 40.00 \\
SpaceR & 60.00 & 36.67 & 67.50 & 80.00 & 57.14 & 20.00 & 22.86 & 25.71 & 43.33 & 56.40 & 60.00 & 18.81 & 26.67 & 66.67 \\
SpaceOm & 52.50 & 43.33 & 52.50 & 88.57 & 31.43 & 25.71 & 25.71 & 77.14 & 33.33 & 32.30 & 65.71 & 5.00 & 40.00 & 53.33 \\
LLaVA-OV-1.5-8B & 57.50 & 63.33 & 7.50 & 88.57 & 42.86 & 31.43 & 31.43 & 45.71 & 33.33 & 32.30 & 54.29 & 26.67 & 30.00 & 80.00 \\
SpatialThinker-3B & 47.50 & 36.67 & 55.00 & 91.43 & 37.14 & 20.00 & 42.86 & 65.71 & 36.67 & 28.30 & 54.29 & 5.77 & 43.33 & 60.00 \\
Qwen3-VL-8B & 55.00 & 66.67 & 37.50 & 80.00 & 22.86 & 25.71 & 22.86 & 77.14 & 33.33 & 39.10 & 68.57 & 40.95 & 16.67 & 36.67 \\
GEM-2B & 65.00 & 40.00 & 47.50 & 74.29 & 54.29 & 22.86 & 42.86 & 54.29 & 33.33 & 22.90 & 62.86 & 27.62 & 16.67 & 46.67 \\
Cambrian-S-7B & 82.50 & 46.67 & 60.00 & 82.86 & 28.57 & 22.86 & 28.57 & 5.71 & 16.67 & 58.90 & 65.71 & 20.00 & 30.00 & 46.67 \\
GLM-4.6V-Flash & 25.00 & 73.33 & 32.50 & 94.29 & 48.57 & 25.71 & 40.00 & 68.57 & 3.33 & 20.20 & 65.71 & 36.27 & 16.67 & 23.33 \\
VLM-3R-7B & 67.50 & 53.33 & 12.50 & 91.43 & 22.86 & 25.71 & 25.71 & 11.43 & 33.33 & 13.50 & 65.71 & 13.64 & 33.33 & 46.67 \\
Spatial-MLLM & 27.50 & 10.00 & 5.00 & 88.57 & 25.71 & 25.71 & 34.29 & 34.29 & 36.67 & 10.80 & 51.43 & 16.19 & 33.33 & 60.00 \\
SpatialRGPT-8B & 62.50 & 16.67 & 5.00 & 71.43 & 25.71 & 37.14 & 31.43 & 17.14 & 23.33 & 10.80 & 54.29 & 8.57 & 16.67 & 63.33 \\
Janus-Pro-7B & 67.50 & 10.00 & 5.00 & 74.29 & 37.14 & 17.14 & 42.86 & 8.57 & 13.33 & 28.20 & 65.71 & 5.71 & 16.67 & 50.00 \\
Janus-1.3B & 52.50 & 56.67 & 0.00 & 48.57 & 54.29 & 22.86 & 37.14 & 0.00 & 6.67 & 28.20 & 60.00 & 0.00 & 16.67 & 16.67 \\
SpatialBot-3B & 57.50 & 10.00 & 5.00 & 60.00 & 37.14 & 31.43 & 40.00 & 22.86 & 33.33 & 10.80 & 51.43 & 9.05 & 6.67 & 16.67 \\
\midrule
\rowcolor{green!5}\multicolumn{15}{l}{\textbf{\textit{Visual Answering}}} \\
GPT Image 2 & 82.50 & 63.33 & 52.50 & 80.00 & 45.71 & 40.00 & 25.71 & 77.14 & 40.00 & 54.07 & 31.43 & 60.40 & 40.00 & 70.00 \\
Nano Banana 2 & 82.50 & 86.67 & 62.50 & 60.00 & 45.71 & 25.71 & 31.43 & 65.71 & 16.67 & 20.95 & 37.14 & 42.50 & 40.00 & 63.33 \\
GPT-5 Image Mini & 52.50 & 73.33 & 77.50 & 54.29 & 42.86 & 28.57 & 28.57 & 34.29 & 10.00 & 30.66 & 40.00 & 14.05 & 36.67 & 36.67 \\
Seedream 4.5 & 30.00 & 90.00 & 40.00 & 45.71 & 45.71 & 20.00 & 25.71 & 51.43 & 10.00 & 36.08 & 40.00 & 9.12 & 36.67 & 53.33 \\
JoyAI-Image & 55.00 & 53.33 & 5.00 & 54.29 & 40.00 & 17.14 & 25.71 & 57.14 & 10.00 & 14.20 & 34.29 & 34.51 & 33.33 & 60.00 \\
FLUX.2 [klein] 9B & 25.00 & 80.00 & 20.00 & 65.71 & 40.00 & 8.57 & 45.71 & 11.43 & 23.33 & 16.50 & 28.57 & 28.61 & 36.67 & 40.00 \\
SenseNova-Vision-7B-MoT & 22.50 & 83.33 & 12.50 & 11.43 & 48.57 & 8.57 & 37.14 & 88.57 & 33.33 & 4.00 & 0.00 & 51.13 & 20.00 & 36.67 \\
FLUX.2 [klein] 4B & 20.00 & 46.67 & 32.50 & 68.57 & 40.00 & 14.29 & 31.43 & 22.86 & 6.67 & 18.53 & 17.14 & 11.79 & 36.67 & 36.67 \\
OmniGen-v1 & 7.50 & 93.33 & 10.00 & 8.57 & 48.57 & 5.71 & 42.86 & 2.86 & 33.33 & 17.16 & 28.57 & 5.63 & 40.00 & 30.00 \\
Qwen-Image-Edit-2511 & 10.00 & 43.33 & 15.00 & 54.29 & 31.43 & 14.29 & 34.29 & 17.14 & 23.33 & 28.59 & 37.14 & 4.74 & 33.33 & 23.33 \\
BAGEL-7B-MoT & 10.00 & 36.67 & 20.00 & 42.86 & 40.00 & 14.29 & 37.14 & 2.86 & 33.33 & 0.00 & 40.00 & 5.63 & 33.33 & 40.00 \\
\midrule
\bottomrule
\end{tabular}

\footnotesize
Cnt.: Counting; Dep.: Depth; Ori.: Orientation; Rel.: Relationship; Pers.: Perspective; Ground.: Spatial Grounding; M-Hop: Multi-hop; Pred.: Prediction; Feas.: Geometric Feasibility; Aff.: Affordance; Nav.: Navigation; Traj.: Trajectory.
\end{center}
\end{landscape}
\clearpage

\clearpage
\end{document}